\documentclass{article} 

\usepackage{arxiv}

\usepackage[utf8]{inputenc} 
\usepackage[T1]{fontenc}    
\usepackage{newtxtext,newtxmath}
\usepackage{graphicx}
\usepackage{subcaption}
\usepackage[backend=biber,style=numeric]{biblatex}
\usepackage{amsmath}
\usepackage{microtype}      
\usepackage{nicefrac}       
\usepackage{booktabs}       
\PassOptionsToPackage{table}{xcolor} 
\usepackage[most]{tcolorbox}

\usepackage{hyperref}       
\usepackage{url}            

\usepackage{amsfonts}       

\usepackage{graphicx}
\graphicspath{{./images/}}
\usepackage{subcaption}

\usepackage{float}
\usepackage{longtable}

\makeatletter
\renewcommand\subsubsubsection{\@startsection{paragraph}{4}{\z@}{-2.5ex\@plus -1ex \@minus -.25ex}{1.25ex \@plus .25ex}{\normalfont\normalsize\bfseries}}
\makeatother

\usepackage{tikz}
\usetikzlibrary{arrows.meta, positioning}

\usepackage[colorinlistoftodos]{todonotes}

\usepackage{listings}
\usepackage{fancyvrb}
\usepackage{fvextra}
\lstdefinelanguage{Cypher}{
  morekeywords={
    MATCH, OPTIONAL, WHERE, WITH, RETURN, DISTINCT, AS, ORDER, BY,
    SKIP, LIMIT, UNWIND, CREATE, MERGE, SET, DELETE, DETACH,
    REMOVE, FOREACH, INDEX, CONSTRAINT, ON, USING, CALL,
    STARTS, ENDS, CONTAINS, CASE, WHEN, THEN, ELSE, END, AND, OR, NOT
  },
  sensitive=true,
  morecomment=[l]{//},         
  morestring=[b]',             
  morestring=[b]"              
}

\DefineVerbatimEnvironment{wrappedverb}{Verbatim}{
  breaklines     = true,      
  breakanywhere  = true,      
  breakindent    = 0pt,       
  breaksymbolleft=\tiny\(\hookleftarrow\) 
}

\newcommand{\code}[1]{\texttt{\detokenize{#1}}}
\newcommand{\result}[1]{\textbf{\textit{#1}}} 

\title{A Human-in-the-Loop, LLM-Centered Architecture for Knowledge-Graph Question Answering}

\author{
 Larissa Pusch \\
  Department of Visual and Data-Centric Computing\\
  Zuse Institute Berlin\\
  Berlin, Germany \\
  \texttt{pusch@zib.de} \\
  \And
Alexandre Courtiol\\
  Department of Evolutionary Genetics\\
  Leibniz Institute for Zoo and Wildlife Research\\
  Berlin, Germany \\
  \texttt{courtiol@izw-berlin.de}
  \And
 Tim Conrad \\
  Department of Visual and Data-Centric Computing\\
  Zuse Institute Berlin\\
  Berlin, Germany \\
  \texttt{conrad@zib.de} \\
}

\begin{document}

\maketitle

\begin{abstract}
Large Language Models (LLMs) excel at language understanding but remain limited in knowledge-intensive domains due to hallucinations, outdated information, and limited explainability. Text-based retrieval-augmented generation (RAG) helps ground model outputs in external sources but struggles with multi-hop reasoning. Knowledge Graphs (KGs), in contrast, support precise, explainable querying, yet require a knowledge of query languages. This work introduces an interactive framework in which LLMs generate and explain \texttt{Cypher} graph queries and users iteratively refine them through natural language. 
Applied to real-world KGs, the framework improves accessibility to complex datasets while preserving factual accuracy and semantic rigor and provides insight into how model performance varies across domains. Our core quantitative evaluation is a 90-query benchmark on a synthetic movie KG that measures query explanation quality and fault detection across multiple LLMs, complemented by two smaller real-life query-generation experiments on a Hyena KG and the MaRDI (Mathematical Research Data Initiative) KG. The code is available at \url{https://git.zib.de/lpusch/talk2cypher}.
\end{abstract}

\clearpage
\section{Introduction}

Large Language Models (LLMs) have become integral to question answering (QA) tasks due to their advanced language understanding and generation capabilities. Yet their utility is often limited in knowledge-intensive domains where precision and interpretability are critical. In such settings, LLMs frequently suffer from outdated or hallucinated knowledge and may generate a plausible-sounding but incomplete or incorrect answer \cite{huang2025survey} without clearly indicating the sources or reasoning steps underlying its conclusion \cite{arrieta2020explainable}.

To address these limitations, LLM-based retrieval-augmented generation (RAG) techniques ground model outputs in external information sources without including these sources in the LLM training process. However, because most RAG systems rely on text-based retrieval \cite{zhao2024retrieval}, they struggle to support multi-hop reasoning - that is, answering questions that require combining information from multiple, indirectly connected sources. For example, determining which researchers cited a paper that implemented a specific algorithm involves chaining through authors, publications, and software entries. Traditional semantic similarity search often fails to capture such indirect relationships, especially when relevant information spans documents or concepts with little lexical overlap \cite{liu2025hoprag}.

Knowledge Graphs (KGs), structured representations of entities and their relationships, offer an alternative. Unlike unstructured text, KGs encode domain knowledge in a way that supports logical inference, schema evolution, and semantic querying \cite{hogan2021knowledge}. This makes them well-suited for applications requiring contextualized, verifiable answers, such as natural sciences, biomedicine or mathematics. Search engines have long recognized these advantages: Google's Knowledge Graph rollout in 2012 improved ranking and disambiguation by resolving queries over graph entities instead of raw keywords \cite{Singhal2012,hogan2021knowledge}.

Despite recent progress, KG–based retrieval-augmented generation (KG-RAG) pipelines can still be suboptimal. Three design choices are particularly limiting:

\begin{enumerate}
\item \textbf{Linearised triples erase graph structure.} Some pipelines flatten a KG or its subgraph into a list of subject–predicate–object triples and pass them to the language model verbatim or in rigid templates \cite{zhu2025graph}. The resulting text is verbose, inflates the context window, and forces the model to reconstruct relational structure that was explicit in the original schema.

\item \textbf{Black-box retrieval hides the reasoning path.} Other systems embed the graph using a graph neural network (GNN) and retrieve nodes or paths by using similarity measures \cite{zhu2025graph}. While the retrieved items and their similarity scores are inspectable, the transformation that maps the query to that vector, and thus the full reasoning path, remains implicit and can only be approximated with post-hoc explainability tools. An explicit \texttt{SPARQL} or \texttt{Cypher} query would expose an auditable, deterministic path from data to answer.

\item \textbf{No user-in-the-loop control.} Many architectures treat the user as a passive consumer: once retrieval and generation start, there is no hook to correct a mistaken entity match, broaden or narrow the scope, or override the model’s drafting choices. Lack of such feedback channels reduces transparency and flexibility and weakens trust.
\end{enumerate}

In our previous work \cite{pusch2024combining}, we evaluated LLMs on translating biomedical questions into \texttt{Cypher} queries over a biomedical KG. We released a benchmark question set and an open-source interface, but the lack of means to debug, refine or explain queries meant that any failure required a complete regeneration. This highlights a disconnect between user intent and graph-based query execution.

To close this gap, we now propose an interactive framework in which LLMs not only generate graph queries but also explain them in natural language. Users can provide feedback through natural-language amendments, which are used to iteratively update the query. This creates a feedback loop that supports interpretability and precise control. The explanation capabilities of the framework are evaluated on a simple movie Knowledge Graph, coupled with a structured evaluation query set. Additionally, we conduct two case studies to evaluate the pipeline on several LLMs and two heterogeneous graphs: a Hyena KG and a mathematical research KG from the Mathematical Research Data Initiative (MaRDI) \cite{schubotz2023bravo}. 

The Hyena KG was derived from a subset of data collected for the long-term monitoring of a free-ranging population of spotted hyenas from the Ngorongoro Crater in Tanzania (see \url{https://hyena-project.com}). Biologists have been studying this population for nearly 30 years, collecting a wide range of data (on demography, behaviour, morphology, physiology, diseases and genetics) to address fundamental ecological and evolutionary questions (e.g. \emph{Why do some individuals disperse to other social groups while others do not?} \cite{davidian2016disperse}), as well as specific questions related to this particular population (e.g., \emph{How does pastoralism impact these hyenas?} \cite{dheer2022pastoralism}). As the composition of social groups (called clans) is dynamic, and interactions between individuals are strongly influenced by relatedness, much of the work on the hyena data requires tracking the movements of individuals across clans and their affiliations. This means that researchers require in-depth knowledge of the data structure and expertise in database manipulation to answer even the simplest questions. In order to reduce the barriers to entry for new researchers, members of the Ngorongoro Hyena Project have developed a substantial \texttt{R} package (\emph{hyenaR}, with over 450 exported functions at the time of writing). While this tool is already helpful, users still need to be proficient in the \texttt{R} language, which not all biologists are. Moreover, new projects tend to require additional functions, thereby increasing the development and maintenance burden for the developers. Finally, a tool such as \emph{hyenaR} cannot easily be adapted for the study of other long-term projects by other research groups since much of the coding is tailored to the idiosyncratic structure of the data. A reliable LLM/KG combination could be a more appealing alternative because it represents a generic solution that would largely remove the programming burden while providing an interface accessible to all.

MaRDI’s open Knowledge Graph already consists of over 700 million triples, exposing them through public \texttt{SPARQL} and REST endpoints. This infrastructure is designed to support FAIR principles in mathematics, enabling users to, for example, trace a theorem to its software implementation or link benchmark datasets to all related publications. However, the technical barrier posed by \texttt{SPARQL} continues to limit accessibility. Our approach, applied to a subgraph of MaRDI, could remove that barrier without sacrificing the precision and interpretability provided by curated KGs.

To compare many LLMs under identical conditions, we keep the question sets for the MaRDI and Hyena KGs small. This trades depth per KG for breadth across models. As a result, the study is optimized to reveal failure modes and inconsistencies across models, rather than to provide an exhaustive ranking of overall performance.

We evaluate our system on the three KGs, measuring the faithfulness of natural-language explanations on the synthetic movie KG and the accuracy of generated queries on the MaRDI and Hyena KG. This work contributes toward making semantic graph querying more transparent, interactive, and accessible to non-experts.

In summary, we designed a system that combines the expressiveness of \texttt{Cypher} with the accessibility of LLMs, explicitly targeting the limitations identified in existing KG-based RAG systems. The key contributions of this work are:

\begin{enumerate}
    \item An interactive natural-language query framework with transparent reasoning that lets users ask questions about the contents of a Knowledge Graph, inspect and refine the generated queries through conversational feedback, and obtain natural-language explanations and plausibility checks that make the reasoning path from data to answer explicit while grounding responses in curated graph data, reducing the risk of hallucinations.
    \item Empirical evaluation on synthetic (movie-domain), and domain-expert (MaRDI KG and Hyena KG) graphs to assess explanation quality, fault detection and query generation.
    \item Insights into common LLM failure modes when generating Cypher queries, performance variability across models, and the importance of measuring both accuracy and amendment efficiency.
\end{enumerate}

The novelty of our approach lies in combining transparent LLM-generated queries with an interactive amendment loop that lets users shape and audit the reasoning process without needing to write \texttt{Cypher} themselves.
In the following, we present the architecture and main components of this framework, demonstrating how they work together to enable query generation, execution, transparent explanation, and iterative refinement. To make the framework’s capabilities more concrete, we begin with an example interaction that illustrates these functionalities. This also serves to highlight the importance of transparency and control in Knowledge Graph interaction.

\subsection{Example Interaction}

Before describing the internal components, we present an example interaction to give a concrete impression of the framework in action. In this two-turn exchange, a user asks a natural-language question, the LLM generates the corresponding graph (\texttt{Cypher}) query, it gets executed on the graph database and results are returned. The example also shows how the framework handles clarification and refinement when the initial interpretation of a query does not align with the user’s intent.

\renewcommand{\arraystretch}{1.2}
\begin{tabular}{>{\columncolor{gray!15}}p{0.08\linewidth} p{0.86\linewidth}}
\textbf{T1} &
\textit{User:} Which authors does \texttt{graphclust} have? \\[2pt]

& \textbf{LLM → Cypher} \\
& \begin{lstlisting}[language=cypher,frame=single]
MATCH (p:Publication {name:"graphclust"})-[:HAS_AUTHOR]->(a:Author)
RETURN a.name;
\end{lstlisting} \\[-6pt]

& \textbf{DB → Result}\result{ \textit{Empty Result} because no
\texttt{Publication} node called *graphclust* exists.} \\[2pt]

& \textbf{LLM (short explanation).} \begin{enumerate}
  \item Locate the single \texttt{Publication} node whose \texttt{name} equals~``graphclust''.
  \item Traverse every outgoing \texttt{:HAS\_AUTHOR} edge to neighbouring \texttt{Author} nodes.
  \item Return the \texttt{name} of each such \texttt{Author}.
\end{enumerate}

\hspace{3ex}\textit{Purpose (one line):}  
List all authors associated with the publication \texttt{graphclust}.

\hspace{3ex}\textit{Plausibility check:}
The pattern is sensible: publications “have” authors and the labels/edge name fit the domain.
Some schemas flip the edge
\(\bigl(\texttt{Author}\xrightarrow{\texttt{:AUTHORED}}\texttt{Publication}\bigr)\),
but the chosen \texttt{Publication$\rightarrow$Author} direction is equally valid. \\

\multicolumn{2}{c}{} \\[1ex]
\arrayrulecolor{black}\specialrule{1.2pt}{0pt}{0pt}
\multicolumn{2}{c}{} \\[1ex]

\textbf{T2} &
\textit{User:} Actually, I meant the \emph{software package}, not the publication. \\[2pt]

& \textbf{LLM → Cypher} \\
& \begin{lstlisting}[language=cypher,frame=single]
MATCH (s:SoftwarePackage {name:'graphclust'})-[:HAS_AUTHOR]->(a:Author)
RETURN a.name;
\end{lstlisting} \\[-6pt]

& \textbf{DB → Result} \result{ Tabea Rebafka} \\[2pt]

\end{tabular}
\label{fig:interaction-flow}

This example highlights the framework’s core capabilities: translating natural language into structured queries, explaining the resulting logic, and enabling users to refine them through conversational feedback. This allows the system to bridge the gap between natural language and formal query syntax, without requiring users to learn a query language, such as \texttt{Cypher}.

These capabilities are supported by a modular architecture which allows for switching of LLMs or KGs. The next section introduces this architecture and describes its core components: query generation, explanation, execution, and refinement.

\section{Framework Architecture}\label{sec:framework}
To overcome the limitations of current KG-based systems, and make graph-structured knowledge more accessible, we developed an interactive framework that allows users to query a given Knowledge Graph using only natural language (NL). At its core is a modular pipeline that translates natural-language user questions into executable queries (Generator), runs them against the underlying graph database (Executor), and returns the results. The same large language model (LLM) is also used to explain the generated queries (Explainer) and to process follow-up instructions for refinement (Amender). Unlike conventional KG-based retrieval systems, which often treat users as passive recipients of generated answers, our framework emphasizes transparency and user control, which enables a conversational interaction with the underlying structured graph data, without requiring users to write \texttt{Cypher} queries.

This paper builds on our earlier work introduced in \cite{pusch2024combining}, introducing and evaluating the missing explanation and amendment capabilities. These enhancements shift the system from a static query generator to an interactive dialogue agent for graph exploration. 

\begin{figure}[H]
    \centering
    \includegraphics[width=0.95\linewidth]{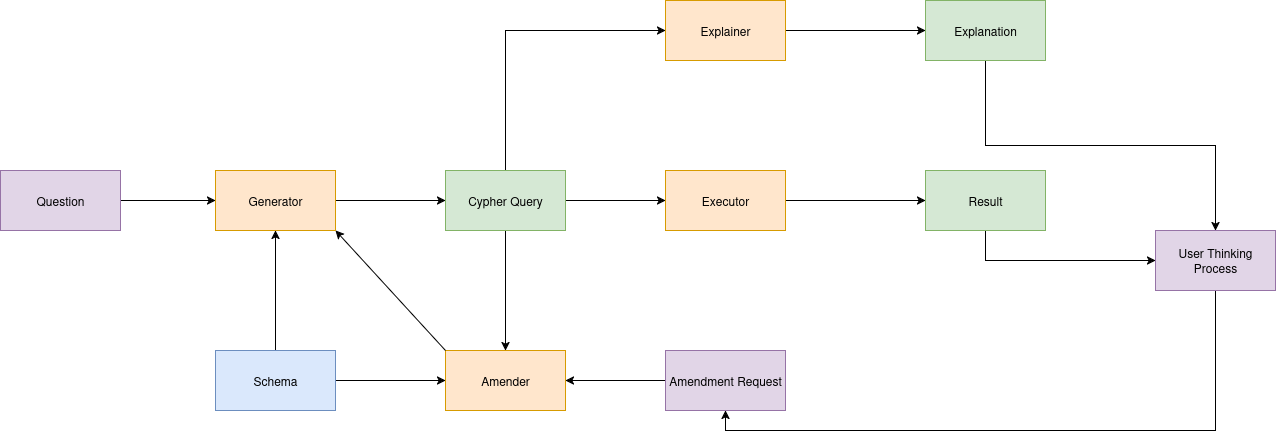}
    \caption{Architecture Diagram; Purple Nodes are contributed by the user, orange nodes are modules, the blue one is the graph schema and the green nodes are pipeline outputs.}
    \label{fig:pipeline}
\end{figure}

\subsection{Framework Main Components}

\subsubsection*{Query Generation}

The query generation component translates a user’s natural-language question into an executable \texttt{Cypher} query. Its primary goal is to preserve semantic intent while conforming to the graph’s structure and constraints.

\noindent \textbf{Inputs and Outputs}: The component takes the user’s question and the Knowledge Graph schema as input. Its output is a cleaned, executable \texttt{Cypher} query string.

\noindent \textbf{Implementation Details}:
The translation is performed using a schema-aware \texttt{LangChain} prompt (\autoref{subsec:gen_prompt}) that constrains vocabulary and structure to valid node and edge types. The LLM (e.g., GPT-4, Claude, or DeepSeek) is wrapped in a \texttt{LangChain}-compatible class, ensuring consistent behavior and easy model switching. Post-processing removes markdown wrappers, tags like \textit{thinking}, and other extraneous content, resulting in a syntactically correct query.

\subsubsection*{Query Execution}

This component connects to a \texttt{Neo4j} instance and executes the \texttt{Cypher} query returned by the generator. It is responsible for retrieving structured answers from the Knowledge Graph.

\noindent \textbf{Inputs and Outputs}:
Its input is the cleaned \texttt{Cypher} query along with a database connector. The output is a list of structured records representing the query result.

\noindent \textbf{Implementation Details}:
Execution is handled using \texttt{LangChain}’s \texttt{Neo4jGraph} object, which provides an abstraction over query submission and result retrieval. Results are returned as a \texttt{Python} data-structure, ready for display or further processing.

\subsubsection*{Query Explanation}
To improve transparency and user trust, this component generates human-readable explanations of the \texttt{Cypher} query. These explanations help users understand how a query works and why a particular result was produced.

\noindent \textbf{Inputs and Outputs}:
It takes a \texttt{Cypher} query as input and returns a structured explanation, including a step-by-step walkthrough, error flagging and a concise summary.

\noindent \textbf{Implementation Details}:
The explanation is generated using the same LLM infrastructure as the query generation component but with a different prompt (\autoref{subsec:explanation_prompt}). The prompt instructs the model to describe the query’s purpose and logic in plain language. Additionally, the model is asked to flag issues such as inverted relationships, invalid node types, or implausible graph patterns. This supports interpretability without requiring users to read \texttt{Cypher} directly.

\subsubsection*{Query Amendment}
The amendment component enables iterative refinement by allowing users to update an existing query using natural language. Rather than regenerating the query from scratch, the system applies targeted modifications based on user feedback.

\noindent \textbf{Inputs and Outputs}:
Inputs include the existing \texttt{Cypher} query, the user’s amendment instruction, and the schema. The output is a revised \texttt{Cypher} query string.

\noindent \textbf{Implementation Details}:
Amendments are handled via a dedicated prompt (\autoref{subsec:amend_prompt}) that asks the LLM to edit the current query in accordance with the new instruction while adhering to schema constraints. This approach aims to avoid overcorrection and retain relevant query structure. The LLM’s output is parsed and returned for immediate re-execution or explanation. This component enables conversational query correction, allowing users to adjust filters, entity types, or traversal logic without starting over.

\subsection{Summary}

We have now described the key components of the framework and how they interact to form an end-to-end pipeline. This allows for the translation of natural-language questions into executable graph queries, which can be run against a knowledge graph, and the results then be returned to the user. More technically, a large language model generates a \texttt{Cypher} query, which is executed via the \texttt{Neo4j} driver, and the resulting records are displayed in the user interface. The same model can also provide a step-by-step natural-language explanation of the query or apply an amendment instruction expressed in natural language, without requiring full regeneration.

This process was illustrated in the “graphclust” example above, where a user refines their query to distinguish between a publication and a software package. The interaction demonstrates how the framework supports clarification and allows non-experts to explore graph-structured data without needing to read or write \texttt{Cypher}.

With the full framework in place, we first focus on our most detailed quantitative experiment, the explanation component in a controlled benchmark, and then turn to two case studies on real-world KGs.

\section{Experiment: How well can LLMs explain \texttt{Cypher} Queries?}\label{sec:exp1}

To ensure that users can trust and control the query process, they need to understand what a generated \texttt{Cypher} query actually does. In this section, we evaluate whether Large Language Models (LLMs) can support that understanding by providing accurate, natural-language explanations and by identifying flawed queries when they occur.

We focus on two main questions:

\begin{enumerate}
    \item Can LLMs generate clear and accurate natural language explanations of \texttt{Cypher} queries?
    \item Can they detect and flag syntactic or semantic problems in queries?
\end{enumerate}

To isolate these abilities from domain complexity, we use a deliberately simple setting: a synthetic Knowledge Graph about movies. This familiar domain makes it easier to spot errors and ensures that poor performance is not due to obscure content or schema.

For instance, a sample query might be:

\begin{lstlisting}[language=cypher]
MATCH (a:Actor)-[:ACTED_IN]->(m:Movie) 
WHERE m.release_year > 2000 
RETURN a.name, m.title
\end{lstlisting}

A correct explanation would state that the query retrieves the names of actors and the titles of movies they acted in, limited to movies released after the year 2000. An inaccurate explanation might omit the year constraint. We deliberately introduced errors into these sample queries, such as replacing the relationship name with one that does not make sense in the context, such as \texttt{(a:Actor)-[EATS]->(m:Movie)}, or more subtle mistakes, such as flipping the direction of the relationship.

With this context in mind, we will next describe the benchmark setup, the models tested, and the evaluation results.

\subsection{Benchmark Dataset}
\label{p:injected_errors}

\begin{figure}[H]
    \centering
    \includegraphics[width=0.5\linewidth]{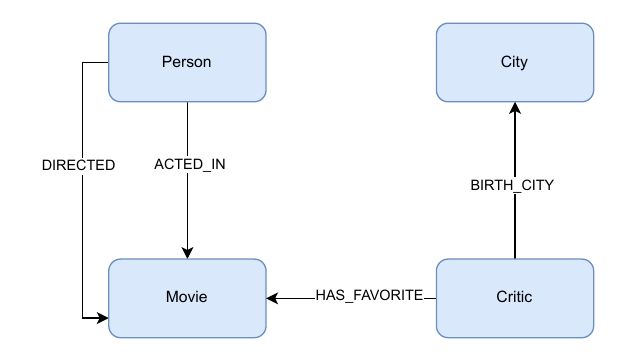}
    \caption{Schema of the synthetic Movie Knowledge Graph created as basis for the queries in the benchmark dataset. The node types are Person, Movie, Critic and City, the relationships are \texttt{DIRECTED}, \texttt{ACTED\_IN}, \texttt{HAS\_FAVORITE} and \texttt{BIRTH\_CITY}.}
    \label{fig:movie_graph}
\end{figure}

We designed a benchmark of 90 \texttt{Cypher} queries (see \autoref{sec:explanation_queries}) set in a movie-domain Knowledge Graph (\autoref{fig:movie_graph}), systematically varying along three dimensions: path lengths, clause types, and error injections.

Unlike existing datasets (\cite{feng2024cypherbench}, \cite{ozsoy2024text2cypher}, \cite{tiwari2024auto}) which lack fine-grained control over query structure and do not contain systematically introduced errors, our benchmark was created from scratch to allow perturbation variation and targeted evaluation.

The resulting query set captures a diverse range of structural and semantic patterns. Below, we illustrate each dimension with examples that show how these variations influence query construction and interpretation.

\subsubsection*{Path length}
\label{subsubsec:n_hops}

To vary structural complexity, queries included 1, 2, or 3 hops, that is, traversal paths of increasing length, where each hop corresponds to following a relationship between connected nodes in the graph.

\begin{tcolorbox}[colback=gray!10, colframe=gray!20, boxrule=0.3pt, arc=2pt, left=2pt, right=2pt, top=2pt, bottom=2pt]
Example:
\vspace{0.2cm}

A 1-hop query might retrieve actors and the movies they acted in:
\begin{lstlisting}[language=cypher]
MATCH (p:Person)-[:ACTED_IN]->(m:Movie) RETURN p.name, m.title
\end{lstlisting}

A 2-hop query could trace from actors to critics via the movie the actor worked in and the critic favored:
\begin{lstlisting}[language=cypher]
MATCH (p:Person)-[:ACTED_IN]->(m:Movie)<-[:HAS_FAVORITE]-(c:Critic) RETURN p.name, c.name
\end{lstlisting}

\end{tcolorbox}

\subsubsection*{Clause composition}
\label{subsubsec:complexity}

To vary logical and syntactic complexity, we used different \texttt{Cypher} constructs: a baseline \texttt{MATCH}; an \texttt{OPTIONAL MATCH} that retrieves additional data when present; two consecutive \texttt{MATCH} clauses connected by \texttt{WITH} to pass intermediate results; a \texttt{WHERE} filter; and a \texttt{CASE} expression that provides if–else logic.

\begin{tcolorbox}[colback=gray!10, colframe=gray!20, boxrule=0.3pt, arc=2pt, left=2pt, right=2pt, top=2pt, bottom=2pt]
Example use of \texttt{CASE}:
\vspace{0.2cm}

\begin{lstlisting}[language=cypher]
MATCH (p:Person {name: "Alice"})-[:ACTED_IN]->(m:Movie) 
RETURN p, m, 
    CASE WHEN m.release_year < 2010 THEN 'old' ELSE 'new' END as movie_age
\end{lstlisting}

\end{tcolorbox}

\subsubsection*{Perturbations} 
\label{subsubsec:perturbation}
To test whether LLMs would recognize perturbations, we deliberately (i) reversed relationship direction, (ii) introduced implausible relationship labels such as EATS (e.g. \texttt{(a:Actor)-[EATS]->(m:Movie)}), (iii) replaced sensible node labels with implausible ones such as Food (e.g. \texttt{(a:Actor)-[ACTED\_IN]->(m:Food)}), (iv) omitted the node type while giving the variable a misleading name, and (v) introduced clause-specific faults: contradictory \texttt{WHERE} constraints (e.g.\ a film released both before and after 2020), illogical \texttt{WHERE} constraints such as a release year of -1, or ill-formed value tests such as \texttt{movie.title > 2020}.

\begin{tcolorbox}[colback=gray!10, colframe=gray!20, boxrule=0.3pt, arc=2pt, left=2pt, right=2pt, top=2pt, bottom=2pt]
Example:
\vspace{0.2cm}

Implausible relationship: The relationship \texttt{EATS} is implausible in this context, testing whether the model recognizes semantic inconsistency.
\begin{lstlisting}[language=cypher]
MATCH (a:Actor)-[:EATS]->(m:Movie)
RETURN a.name, m.title
\end{lstlisting}

\end{tcolorbox}

These variations allow us to assess not just surface-level understanding, but whether LLMs can handle structural, semantic, and logical correctness in \texttt{Cypher} queries.

\subsection{LLMs}

The LLMs tested were selected to be a good representation of the most commonly used models at the time of the experiment that were also easily accessible via \texttt{LangChain}. This includes two GPT variants, a Claude model, and two DeepSeek models. The 70b DeepSeek model was run locally, all other models were accessed via official APIs. 
\begin{table}[H]
\centering
\caption{\label{tab:widgets}LLMs used in explanation trials}
\begin{tabular}{|l|l|}
\hline
llm & version \\\hline
deepseek-r1:70b &  0c1615a8ca32\\
o1-preview-2024-09-12 & 2024-09-12\\
o3-mini-2025-01-31 & 2025-01-31\\
claude-3-7-sonnet-20250219 & 20250219\\
deepseek-reasoner-api & date: 05.03.2025\\
\hline
\end{tabular}
\end{table}

\subsection{Experimental Design} 
We provided the LLM with each query from the benchmark and an instruction prompt (\autoref{subsec:explanation_prompt}) to produce a step-by-step explanation and give a brief summary of the query’s overall purpose. The LLM was also asked to flag errors, with flipped relationships and ill-formed node or relationship names being specifically mentioned in the prompt. Outputs were then manually evaluated by the first author (L.P.) for explanation correctness of the summary; whether errors, when present, were explicitly identified and no errors were flagged if there were none; and whether the summary omitted the year for queries that specified one.
The following example illustrates how the step-by-step summary and the one-sentence-summary were checked for completeness.

\begin{tcolorbox}[colback=gray!10, colframe=gray!20, boxrule=0.3pt, arc=2pt, left=2pt, right=2pt, top=2pt, bottom=2pt]
\begin{Verbatim}[commandchars=|\{\}]
Example Query:
MATCH (|textcolor{red}{p:Person} |textcolor{blue}{name: "Alice"})|textcolor{olive}{-[:ACTED_IN]->}|textcolor{cyan}{(m:Movie)} |textcolor{purple}{RETURN p, m}
Summary:
1. The query finds a |textcolor{red}{node labeled Person} with |textcolor{blue}{the property name set to "Alice"}.  
2. It then follows |textcolor{olive}{outgoing ACTED_IN relationships} from this Person node to find 
   |textcolor{cyan}{connected nodes labeled Movie}.  
3. Finally, it |textcolor{purple}{returns both the Person and the associated Movie nodes}.

Overall, this query |textcolor{purple}{retrieves} the |textcolor{red}{person} |textcolor{blue}{"Alice"} and the |textcolor{cyan}{movies} |textcolor{olive}{in which she has acted}. 
Everything in the query, including the relationship direction and names, is appropriate 
and makes sense.
\end{Verbatim}
\end{tcolorbox}

\subsection{Statistical Analysis}
We computed the mean accuracies and associated 95-percent Wilson confidence intervals of the one-sentence summaries, the detection of injected perturbations, and the avoidance of false positives. For one-sentence summaries, a prediction was counted as correct when the summary provided an accurate and complete description of the query. For problem detection, a success was recorded when the model explicitly identified the perturbation in the queries that contained one. For false positives, a success required that the model did not flag any error in queries without perturbations.

We compared the performances for each of these outcomes of each possible pair of LLMs, out of the five LLMs considered, implementing the McNemar's test for paired binary outcomes in \texttt{Python}. For each pair, we ran the test on a 2x2 contingency table of query-level correctness (correct vs incorrect) and based the test exclusively on the discordant pairs, that is, queries answered correctly by one model and incorrectly by the other. The exact two-sided p-value was obtained from a binomial test on the number of discordant pairs, under the null hypothesis that both models have equal probability of being correct for any query (p = 0.5). To account for multiple pairwise comparisons among models, the p-values were adjusted using the Holm step-down procedure to control the family-wise error rate at a nominal level of 0.05. In addition, we recorded the direction of the difference for each pair of models by comparing the counts of queries where only the row model was correct vs. only the column model was correct, so that adjusted p-values could be interpreted together with the sign of the performance difference.

To examine how the accuracy of one sentence summaries varied across multiple dimensions (as described in \autoref{subsubsec:n_hops}, \autoref{subsubsec:complexity}, \autoref{subsubsec:perturbation}), we used the \texttt{R} package \texttt{spaMM} \cite{rousset2014testing} to fit a series of Generalized Linear Mixed-effects Models (GLMMs) with a binomial link. We generated two alternative response variables representing accuracy: the first considered a strict correctness criterion, and the second considered a relaxed criterion that did not treat a missing year in the one-sentence summary as an error if that year appeared in the step-by-step explanation. In both specifications, the model included LLM, hop count, clause type, and perturbation type as fixed-effect predictors, along with a random intercept for the query. For each GLMM, we conducted Tukey-adjusted post-hoc pairwise comparisons for all multi-level predictors using the function \texttt{glht()} from the package \texttt{multcomp} \cite{multcomp}. To visualize the resulting effects, we computed partial dependence estimates and corresponding confidence intervals using the function \texttt{pdep\_effects(..., intervals = "fixefVar")} and plotted those together with compact letter displays indicating groups not significantly different from each others.
\subsection{Results} 

\subsubsection{Accuracy of one-sentence summaries}

\begin{figure}[htbp]
  \centering
 \resizebox{0.93\textwidth}{!}{%
  \begin{minipage}{\textwidth}
  \begin{subfigure}{0.32\textwidth}
    \centering
    \caption{LLM}
    \includegraphics[width=\linewidth]{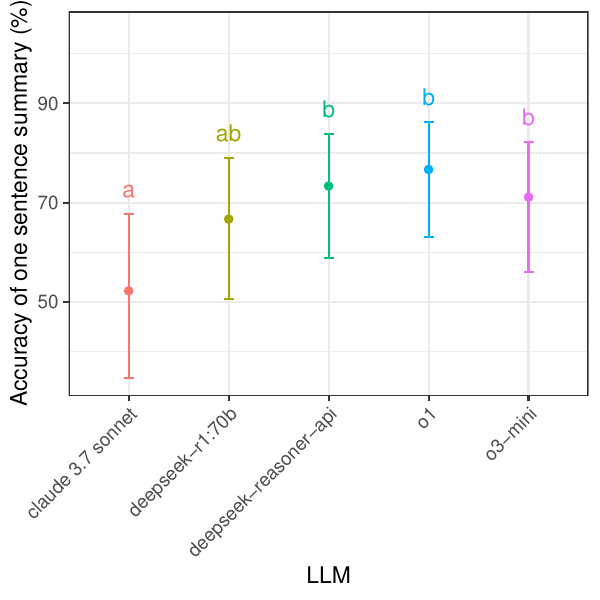}
  \end{subfigure}
  \hfill
  \begin{subfigure}{0.32\textwidth}
    \centering
    \caption{LLM (year not error)}
    \includegraphics[width=\linewidth]{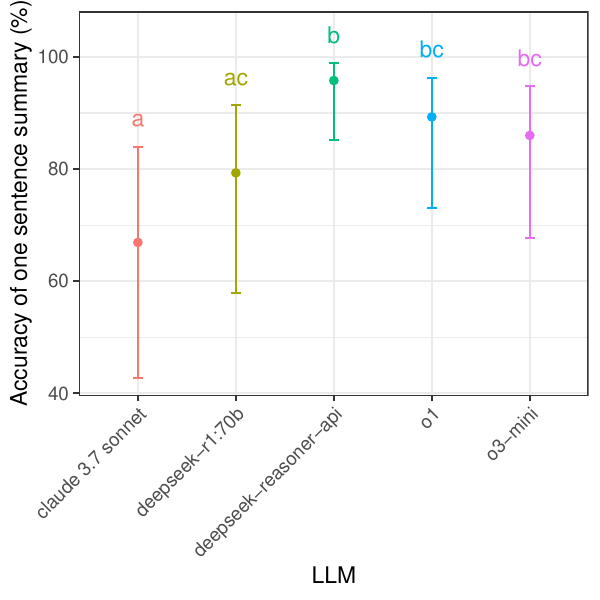}
  \end{subfigure}

  \begin{subfigure}{0.32\textwidth}
    \centering
    \caption{Hop count}
    \includegraphics[width=\linewidth]{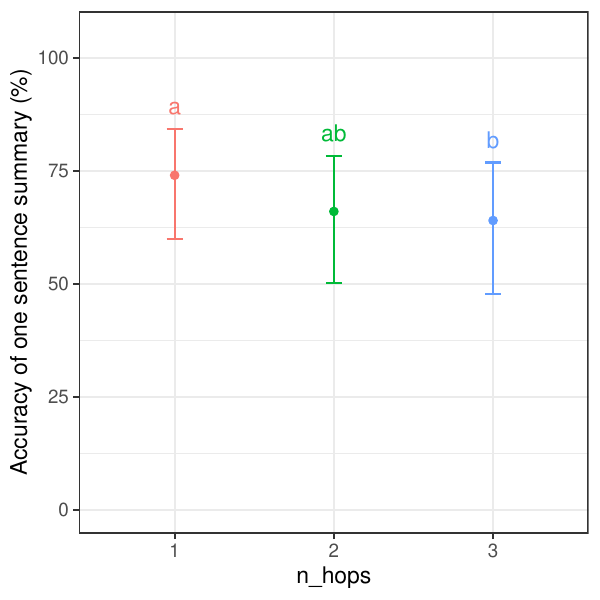}
  \end{subfigure}
  \hfill
  \begin{subfigure}{0.32\textwidth}
    \centering
    \caption{Hop count (year not error)}
    \includegraphics[width=\linewidth]{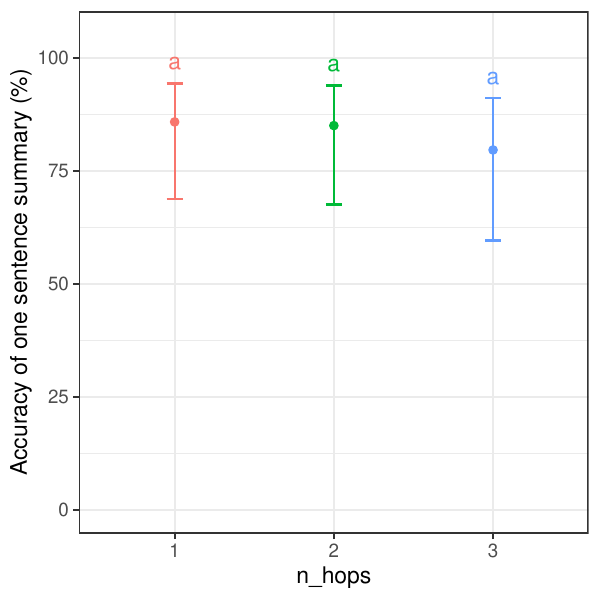}
  \end{subfigure}

  \begin{subfigure}{0.32\textwidth}
    \centering
    \caption{Perturbation}
    \includegraphics[width=\linewidth]{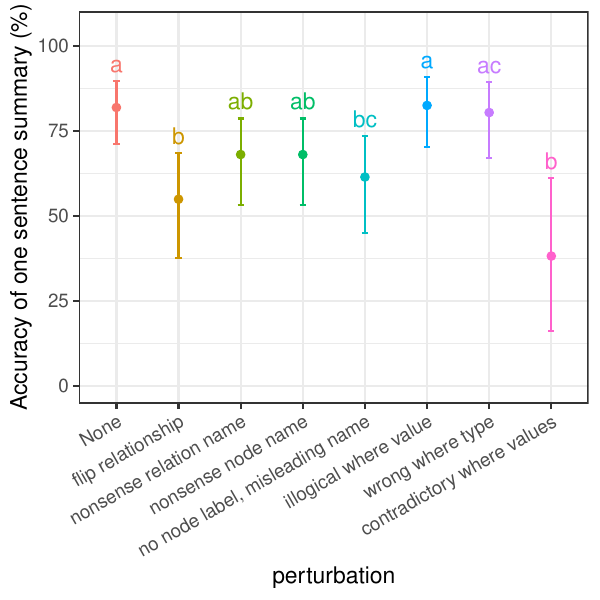}
  \end{subfigure}
  \hfill
  \begin{subfigure}{0.32\textwidth}
    \centering
    \caption{Perturbation (year not error)}
    \includegraphics[width=\linewidth]{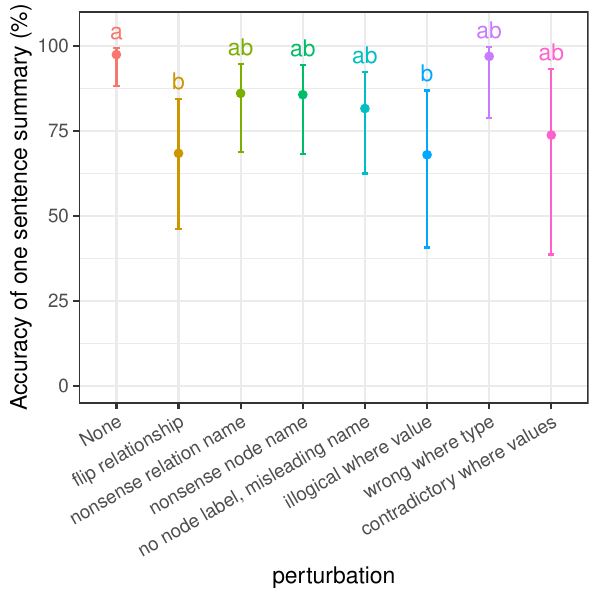}
  \end{subfigure}

  \begin{subfigure}{0.32\textwidth}
    \centering
    \caption{Clause Type}
    \includegraphics[width=\linewidth]{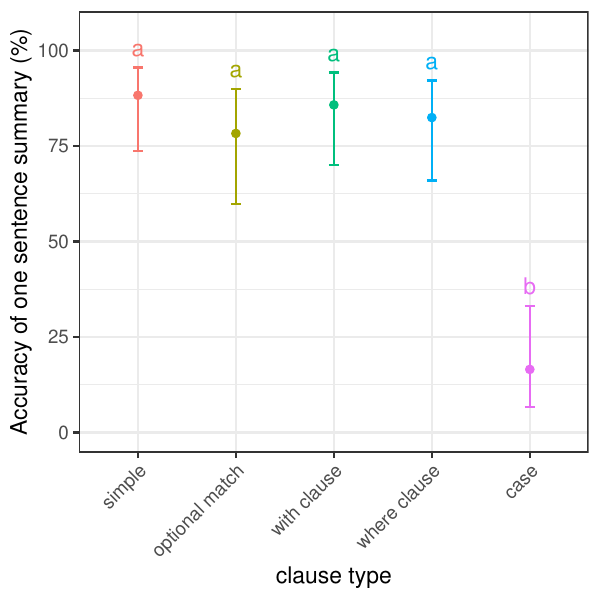}
  \end{subfigure}
  \hfill
  \begin{subfigure}{0.32\textwidth}
    \centering
    \caption{Clause Type (year not error)}
    \includegraphics[width=\linewidth]{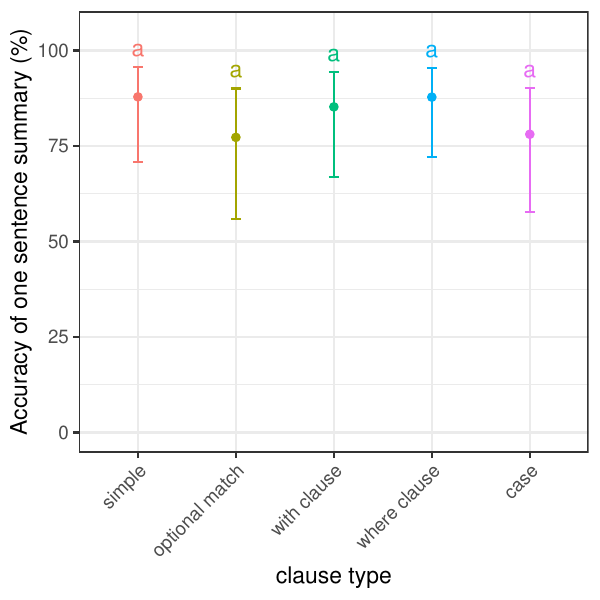}
  \end{subfigure}
    \end{minipage}%
  }
  \caption{Effects on explanation accuracy. Left column: year mismatches counted as errors (strict criterion). Right column: year mismatches not counted as errors (relaxed criterion). These figures were constructed using an additive GLM. If items share a letter, they are not significantly different from each other.}
  \label{fig:dimension_influence}
\end{figure}

\paragraph{Effect of the LMM} The accuracy of the LLMs on correct and complete one-sentence explanation summaries (\autoref{tab:explanation_complete}) exceeded 70\% for o1-preview, deepseek-reasoner-api, and o3-mini, with o1-preview performing the best. Deepseek-r1:70b reached 66\%, and claude 3.7 sonnet was the lowest at 52\%. Trivially wrong statements (e.g., “X is misspelled” followed by a suggestion to change the word to one of identical spelling) and stylistic complaints about node and relationship names were ignored in the scoring because they are unlikely to mislead readers. A Holm-corrected pairwise McNemar test (\autoref{tab:mc_nemar_true_summary}) showed that deepseek-reasoner-api, o1-preview, and o3-mini performed significantly better than claude 3.7 sonnet on one-sentence summaries. This result was confirmed by a GLMM fit controlling for other sources of variation (\autoref{fig:dimension_influence}a \& \autoref{fig:dimension_influence}b). The pattern held irrespective of whether or not year omissions were counted as errors, but deepseek-reasoner-api significantly outperformed deepseek-r1:70b only in the latter case. A detailed breakdown of disagreement counts is provided in \autoref{tab:explanation_first_detailed}.

\begin{figure}[H]
  \centering
  \begin{subfigure}[b]{0.32\linewidth}
    \centering
    \includegraphics[width=\linewidth]{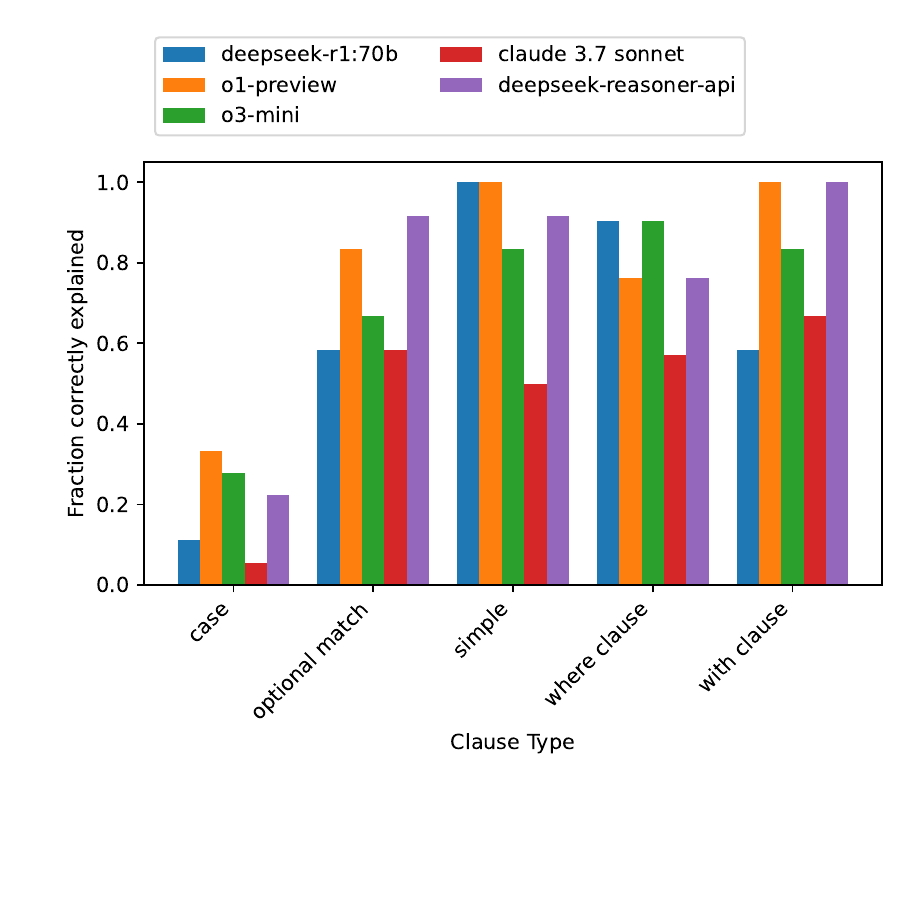}
    \caption{By clause}
    \label{fig:correct_by_clause}
  \end{subfigure}
  \hfill
  \begin{subfigure}[b]{0.32\linewidth}
    \centering
    \includegraphics[width=\linewidth]{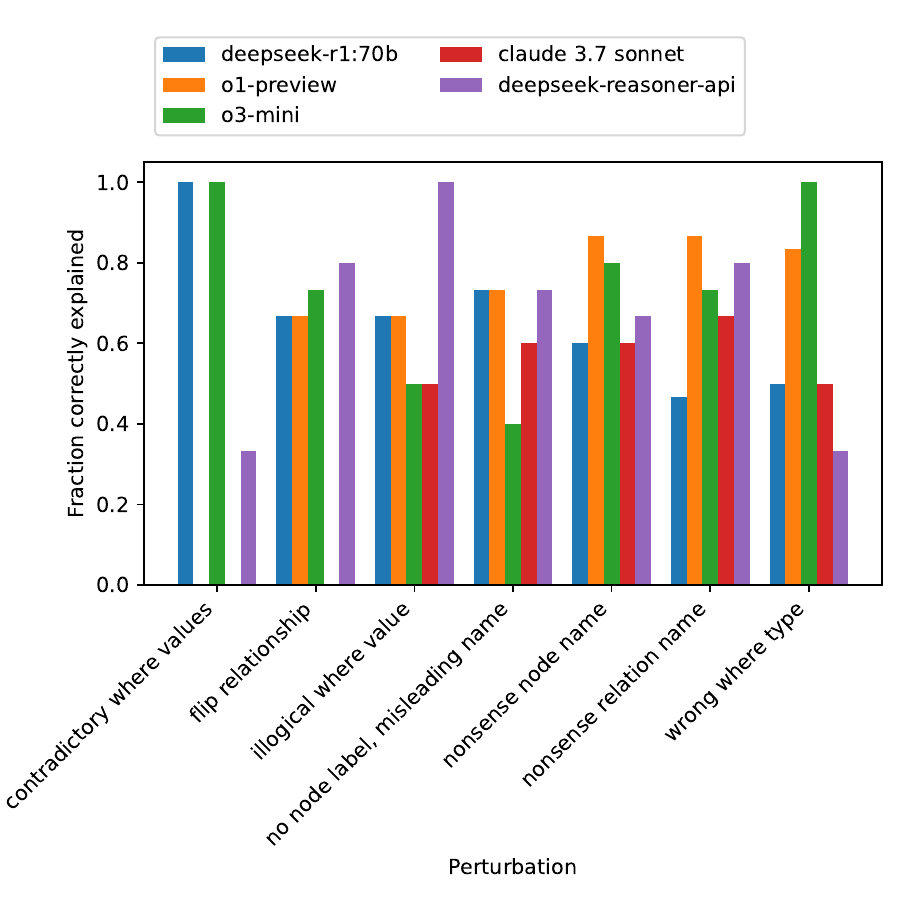}
    \caption{By perturbation}
    \label{fig:correct_by_perturbation}
  \end{subfigure}
  \hfill
  \begin{subfigure}[b]{0.32\linewidth}
    \centering
    \includegraphics[width=\linewidth]{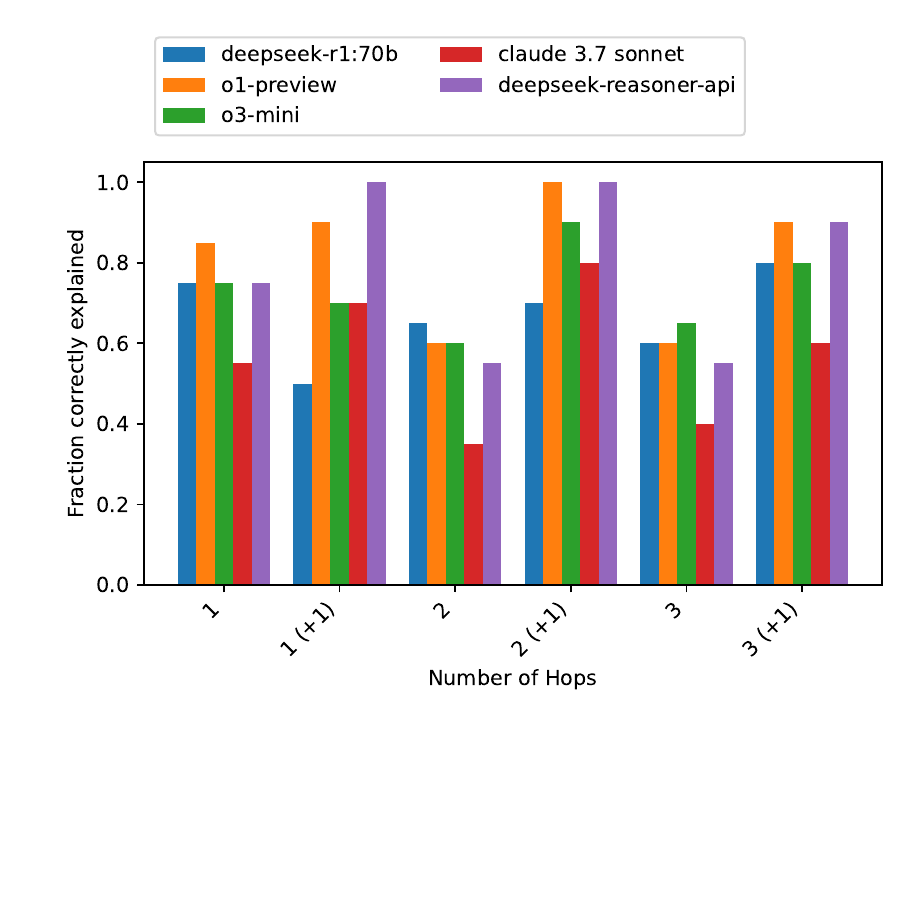}
    \caption{By number of hops}
    \label{fig:correct_by_nhops}
  \end{subfigure}

  \caption{How correctness of one-sentence summaries is influenced by query features.}
  \label{fig:correctness_analysis}
\end{figure}

\paragraph{Effect of query features}
Our query set varied along three axes: clause type, perturbation, and hop count, each potentially affecting explanation performance (\autoref{fig:correctness_analysis}). For clause type ((\autoref{fig:correct_by_clause})), \texttt{CASE} statements stand out as problematic; however, \texttt{CASE} queries uniformly include a year and some LLMs struggled to include missing years in the one-sentence summaries, which likely confounds this result. The GLMM results show that clause type no longer significantly influenced the accuracy of one-sentence summary once year mistakes were overlooked (\autoref{fig:dimension_influence}h).

With respect to perturbations, only \textit{contradictory \texttt{WHERE} values} and \textit{flip relationship} induce complete failures in some LLMs (\autoref{fig:correct_by_perturbation}), resulting in a significantly worse accuracy of one-sentence summary when either of these two perturbations are present, compared to no perturbation (\autoref{fig:dimension_influence}e \& \autoref{fig:dimension_influence}f). More specifically, Claude failed on both for all instances, and o1-preview only failed on all \textit{contradictory \texttt{WHERE} values}. Note that the latter comprises only three queries because it applies solely to \texttt{WHERE} clauses. For other types of perturbation, no cross-model pattern emerged consistently.

Regarding hop count, 1-hop queries were processed more accurately by LLMs than 2- or 3-hop queries (\autoref{fig:correct_by_nhops}). The GLMMs shows that, as for clause type, the hop count no longer significantly influenced the accuracy of one-sentence summary once year mistakes were overlooked (\autoref{fig:dimension_influence}d). When \texttt{OPTIONAL MATCH} or \texttt{WITH} were used to introduce an extra hop ("(+1)" category) to the base hop count, the accuracy surprisingly improved, although such queries were too few to test the resulting differences using a GLMM.

\subsubsection{Ability to flag faults in perturbed queries}

\paragraph{Effect of the LLM} Among perturbed queries, o1-preview, claude 3.7 sonnet, and deepseek-reasoner-api exceeded 80\%, o3-mini scored 77\%, and deepseek-r1:70b scored 68\%  when flagging them as faulty (\autoref{tab:explanation_complete}).
A Holm-corrected pairwise McNemar test (\autoref{tab:mc_nemar_error_flagging}) showed that deepseek-reasoner-api and o1-preview performed significantly better than deepseek-r1:70b. A detailed breakdown of disagreement counts is provided in \autoref{tab:explanation_detection_detailed}.

\begin{figure}[H]
    \centering
    \includegraphics[width=0.5\linewidth]{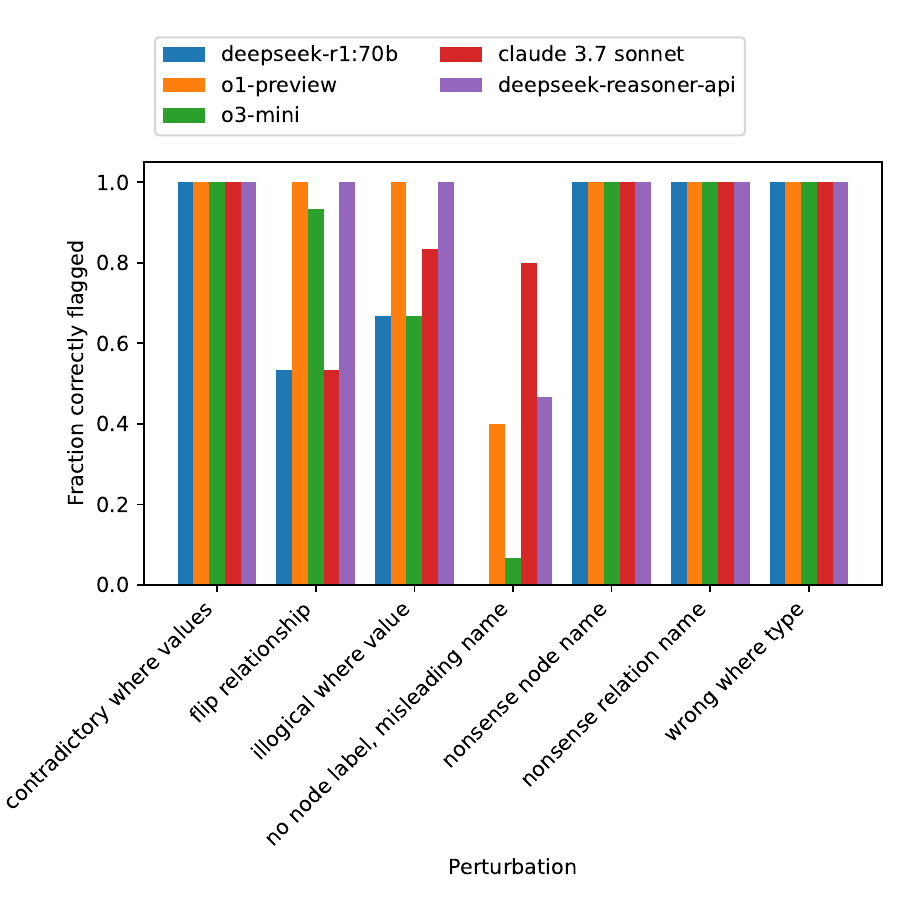}
    \caption{For each perturbation category (definitions in \autoref{p:injected_errors}), the bars show the fraction of queries in which each evaluated LLM correctly signalled that something was wrong.}
    \label{fig:pert}
\end{figure}

\paragraph{Sensitivity to perturbation type in perturbation flagging} Flipped relationships, nonsense node labels and nonsense relationship labels were specifically mentioned in the prompt. All models flagged contradictory \texttt{WHERE} clauses, nonsense node- and relationship names, and type mismatches (\autoref{fig:pert}). Many struggled with the combination of missing node type and misleading node names, where only claude exceeded 50\% at 79\%. For flipped relationship directions, o1-preview and deepseek-reasoner-api were flawless, whereas the others made occasional to frequent mistakes.

\subsubsection{Avoidance of false positives}

\paragraph{Effect of LLM} Deepseek-r1:70b flagged a correct query as defective in 47\% of cases, claude 3.7 sonnet in 33\%, o1-preview in 7\%, and o3-mini and deepseek-reasoner-api never did so (\autoref{tab:explanation_complete}). A Holm-corrected pairwise McNemar test (\autoref{tab:mc_nemar_false_positives}) showed no statistically significant differences between the models. Given the small sample (n=15) and limited power, we interpret differences in this task primarily based on descriptive statistics. A detailed breakdown of disagreement counts is provided in \autoref{tab:false_positive_detailed}.

\begin{table}[h]
\centering
\caption{Accuracy with 95\% Wilson CI for correct one-sentence summary generation, problem detection, and false positives.}
\label{tab:explanation_complete}
\small
\setlength{\tabcolsep}{4pt}
\renewcommand{\arraystretch}{1.1}

\resizebox{\textwidth}{!}{%
\begin{tabular}{l
                r r r l
                r r r l
                r r r l
                r r r l}
\toprule
 & \multicolumn{4}{c}{One-sentence summary} &
   \multicolumn{4}{c}{Problem detection} &
   \multicolumn{4}{c}{False positives} \\
\cmidrule(lr){2-5}\cmidrule(lr){6-9}\cmidrule(lr){10-13}\cmidrule(lr){14-17}
LLM &
  $n$ & correct & acc. & 95\% CI &
  $n$ & correct & acc. & 95\% CI &
  $n$ & correct & acc. & 95\% CI \\
\midrule
claude 3.7 sonnet &
  90 & 47 & 0.522 & [0.420, 0.622] &
  75 & 64 & 0.853 & [0.756, 0.916] &
  15 & 10 & 0.666 & [0.417, 0.848] \\
deepseek-reasoner-api &
  90 & 66 & 0.733 & [0.634, 0.814] &
  75 & 67 & 0.893 & [0.803, 0.945] &
  15 & 15 & 1.000 & [0.796, 1.000] \\
deepseek-r1:70b &
  90 & 60 & 0.666 & [0.564, 0.755] &
  75 & 51 & 0.680 & [0.568, 0.775] &
  15 & 8  & 0.533 & [0.301, 0.752] \\
o1-preview &
  90 & 69 & 0.766 & [0.669, 0.842] &
  75 & 66 & 0.880 & [0.787, 0.936] &
  15 & 14 & 0.933 & [0.702, 0.988] \\
o3-mini &
  90 & 64 & 0.711 & [0.610, 0.795] &
  75 & 58 & 0.773 & [0.667, 0.853] &
  15 & 15 & 1.000 & [0.796, 1.000] \\
\bottomrule
\end{tabular}%
}
\end{table}

\begin{table}[t]
  \centering
  \small
  \setlength{\tabcolsep}{4pt}
  \renewcommand{\arraystretch}{1.1}
  \caption{Pairwise McNemar tests comparing LLM accuracy of one-sentence summary generated. P-values are Holm corrected. Positive values indicate better performance of the model in the row over the model in the column.}
  \label{tab:mc_nemar_true_summary}
  \begin{tabular}{lrrrrr}
    \toprule
    & claude 3.7 sonnet & deepseek-reasoner-api & deepseek-r1:70b & o1-preview & o3-mini \\
    \midrule
    claude 3.7 sonnet &
      - &
      \cellcolor{red!15}\textbf{-0.0119} &
      \cellcolor{red!15}-0.2061 &
      \cellcolor{red!15}\textbf{-0.0000} &
      \cellcolor{red!15}\textbf{-0.0121} \\
    deepseek-reasoner-api &
      \cellcolor{green!15}\textbf{0.0119} &
      - &
      \cellcolor{green!15}1.0000 &
      \cellcolor{red!15}-1.0000 &
      \cellcolor{green!15}1.0000 \\
    deepseek-r1:70b &
      \cellcolor{green!15}0.2061 &
      \cellcolor{red!15}-1.0000 &
      - &
      \cellcolor{red!15}-0.5588 &
      \cellcolor{red!15}-1.0000 \\
    o1-preview &
      \textbf{0.0000} &
      \cellcolor{green!15}1.0000 &
      \cellcolor{green!15}0.5588 &
      - &
      \cellcolor{green!15}1.0000 \\
    o3-mini &
      \cellcolor{green!15}\textbf{0.0121} &
      \cellcolor{red!15}-1.0000 &
      \cellcolor{green!15}1.0000 &
      \cellcolor{red!15}-1.0000 &
      - \\
    \bottomrule
  \end{tabular}
\end{table}

\begin{table}[t]
  \centering
  \small
  \setlength{\tabcolsep}{4pt}
  \renewcommand{\arraystretch}{1.1}
  \caption{Pairwise McNemar tests comparing LLMs on error detection capabilities. P-values are Holm corrected. Positive values indicate better performance of the model in the row over the model in the column.}
  \label{tab:mc_nemar_error_flagging}
  \begin{tabular}{lrrrrr}
    \toprule
    & claude 3.7 sonnet & deepseek-reasoner-api & deepseek-r1:70b & o1-preview & o3-mini \\
    \midrule
    claude 3.7 sonnet &
      - &
      \cellcolor{red!15}-1.0000 &
      \cellcolor{green!15}0.0504 &
      \cellcolor{red!15}-1.0000 &
      \cellcolor{green!15}1.0000 \\
    deepseek-reasoner-api &
      \cellcolor{green!15}1.0000 &
      - &
      \cellcolor{green!15}\textbf{0.0003} &
      \cellcolor{green!15}1.0000 &
      \cellcolor{green!15}0.0312 \\
    deepseek-r1:70b &
      \cellcolor{red!15}-0.0504 &
      \cellcolor{red!15}\textbf{-0.0003} &
      - &
      \cellcolor{red!15}\textbf{-0.0005} &
      \cellcolor{red!15}-0.1953 \\
    o1-preview &
      \cellcolor{green!15}1.0000 &
      \cellcolor{red!15}-1.0000 &
      \cellcolor{green!15}\textbf{0.0005} &
      - &
      \cellcolor{green!15}0.0504 \\
    o3-mini &
      \cellcolor{red!15}-1.0000 &
      \cellcolor{red!15}-0.0312 &
      \cellcolor{green!15}0.1953 &
      \cellcolor{red!15}-0.0504 &
      - \\
    \bottomrule
  \end{tabular}
\end{table}

\begin{table}[t]
\centering
\small
\setlength{\tabcolsep}{4pt}
\renewcommand{\arraystretch}{1.1}
\caption{Pairwise McNemar tests comparing LLMs on false positives. P-values are Holm corrected. Positive values indicate better performance (=less false positives) of the model in the row over the model in the column.}
\begin{tabular}{lrrrrr}
\toprule
 & claude 3.7 sonnet & deepseek-reasoner-api & deepseek-r1:70b & o1-preview & o3-mini \\
\midrule
claude 3.7 sonnet &
  - &
  \cellcolor{red!15}-0.3750 &
  \cellcolor{green!15}1.0000 &
  \cellcolor{red!15}-0.5000 &
  \cellcolor{red!15}-0.3750 \\

deepseek-reasoner-api &
  \cellcolor{green!15}0.3750 &
  - &
  \cellcolor{green!15}0.1406 &
  \cellcolor{green!15}1.0000 &
  -- \\

deepseek-r1:70b &
  \cellcolor{red!15}-1.0000 &
  \cellcolor{red!15}-0.1406 &
  - &
  \cellcolor{red!15}-0.2188 &
  \cellcolor{red!15}-0.1406 \\

o1-preview &
  \cellcolor{green!15}0.5000 &
  \cellcolor{red!15}-1.0000 &
  \cellcolor{green!15}0.2188 &
  - &
  \cellcolor{red!15}-1.0000 \\

o3-mini &
  \cellcolor{green!15}0.3750 &
  -- &
  \cellcolor{green!15}0.1406 &
  \cellcolor{green!15}1.0000 &
  - \\
\bottomrule
\end{tabular}
\label{tab:mc_nemar_false_positives}
\end{table}

\subsection{Discussion}

Across the 90 \texttt{Cypher} queries in our benchmark, all LLMs produced correct and complete one-sentence summaries for more than half of the cases. Only three models, o1-preview, deepseek-reasoner-api, and o3-mini, exceeded the 70\% accuracy mark, which is encouraging but still insufficient for robust real-world use. The gap between the best performer (o1-preview) and the worst (claude) was nearly 20\%, highlighting substantial variation in explanation quality across models. These differences were confirmed by the generalized linear mixed model analysis, which showed significant overall accuracy differences across LLMs, with claude 3.7 sonnet performing reliably worse than deepseek-reasoner-api, o1-preview, and o3-mini under both strict and relaxed evaluation criteria, and deepseek-reasoner-api further outperforming deepseek-r1:70b when year omissions are disregarded.

One weakness recurred throughout the evaluation: omission of years from the one-sentence summaries. For deepseek-reasoner-api, more than four out of five incorrect summaries were wrong for this reason; for o1-preview and o3-mini, the omission accounted for more than half of their errors. This suggests that the summary-generation objective may overweight brevity relative to faithfulness. Explicitly requiring the retention of temporal information during instruction tuning could address this issue and improve performance. The mixed-effects analysis reinforces this interpretation: one-hop queries only outperform three-hop queries when year omissions are counted as errors, indicating that the apparent structural difficulty is largely driven by temporal information loss rather than query length alone.

When detecting flawed queries, models varied in how they balanced sensitivity and specificity. Claude, for example, exhibited high sensitivity (identified many injected defects) but low specificity (raised a relatively high number of false alarms) suggesting a trade-off between the two. In contrast, o1-preview and deepseek-reasoner-api combined high defect-detection rates (above 85\%) with low false-positive rates (below 10\%), showing fewer signs of this trade-off. o3-mini achieved a slightly lower detection rate (around 77\%) but did not produce any false positives. These patterns provide a useful basis for choosing models according to the preferred balance between the two metrics.

Across models, there is no simple one-to-one link between strong perturbation detection and low false positive rates. Most systems that rank highly in detection accuracy also avoid many false positives, but claude 3.7 sonnet breaks this pattern: despite being among the strongest at identifying actual perturbations, its false positive rate was only midrange. Deepseek-reasoner-api and o1-preview illustrate the cleaner relationship, pairing high detection accuracy with near-perfect avoidance of false positives, with o3-mini not far behind. In contrast, deepseek-r1:70b performed worse on both fronts. Overall, while good detection often aligned with better false positive avoidance, the differences among models - especially claude 3.7 sonnet's mix of strong detection and moderate false positive control - show that these abilities do not always improve together. Consistent with these observations, the GLMM shows that under strict evaluation, flipped and contradictory relationships significantly reduce accuracy compared to no perturbation or incorrect WHERE clauses, whereas under the relaxed criterion these differences largely disappear except for the no-perturbation baseline.

Analysing how query features shape performance reveals a small set of consistent patterns. The apparent weakness on \texttt{CASE} clauses is largely a temporality confound: all \texttt{CASE} queries contained years, and missing-year omissions account for a substantial share of errors. The mixed-effects results align with this explanation: CASE queries perform significantly worse than all other clause complexities only when year omissions are treated as errors, further confirming that temporality rather than clause structure drives this effect. The two perturbations that occasionally produced complete failures, contradictory \texttt{WHERE} values and flipped relationship directions, point to specific blind spots (logical consistency checks and small changes with large impact), although flipped relationship directions were specifically mentioned in the prompt \autoref{subsec:explanation_prompt}. Given the small size of the contradictory-\texttt{WHERE} set (n=3), we treat claims here as provisional, especially given that other queries with similar logical consistency checks, such as illogical \texttt{WHERE} values, did not show this pattern. Structurally, base 1-hop queries were easier than 2–3 hops, whereas “(+1)” variants created by \texttt{OPTIONAL MATCH} or \texttt{WITH} were unexpectedly easier than their bases, with the cause being uncertain and meriting follow-up investigation. These patterns were consistent with the effects estimated in our generalized linear mixed models, which showed significant differences across LLMs, hop counts, perturbation types, and clause complexities.

These observations need to be interpreted in light of certain limitations. The scope of this experiment is intentionally narrow, focusing on 90 handcrafted queries in a movie-style Knowledge Graph. While this controlled setting is well-suited for isolating specific behaviours and testing fundamental capabilities, real-world graph queries, especially in domains such as mathematics or biology, can be more complex and embedded in richer schema contexts. Furthermore, the scoring process relied on human judgement, introducing some subjectivity in determining what constitutes a "complete" and correct summary, which is unavoidable given the tool's human-in-the-loop design. Finally, because LLMs evolve rapidly, the snapshot presented here does not offer any guarantees about future capabilities, and subsequent models may surpass the performance observed in this study. Despite these constraints, the evaluation highlights concrete strengths and weaknesses that can directly inform the explanation framework’s future design goals. The following two experiments examine different aspects of the framework, using examples from mathematical (MaRDI) and biological (a population of spotted hyenas) domains to illustrate its capabilities in handling question answering on diverse Knowledge Graphs.

\section{Real-Life Examples: Question Answering on Biological and Mathematical Knowledge Graphs}
To determine how well \texttt{Cypher} query generation from natural-language questions, explained in \autoref{sec:framework}, performs on real-world Knowledge Graphs, we conducted two experiments in distinct domains. The first uses a software-centric slice of the MaRDI Knowledge Graph (including CRAN packages, Zenodo software, cited papers, and authors with zbMATH IDs). The second uses a newly constructed Knowledge Graph derived from long-term field data on a free-ranging population of spotted hyenas in the Ngorongoro Crater, Tanzania (see \url{https://hyena-project.com}), with questions formulated by project experts. Across both experiments, we tested 14 \texttt{LangChain}-compatible LLMs under identical constraints and kept the number of questions deliberately small to enable systematic comparison across many models rather than exhaustive coverage of each graph. Accordingly, the results are not intended as a full benchmark ranking; instead, they should be interpreted as comparative performance under controlled conditions, highlighting failure modes and model-specific difficulties. Our central research question in both cases is whether an LLM can produce a correct query after an initial \texttt{Cypher} generation step and at most two user-submitted amendments, and whether particular questions pose more difficulties than others.

\subsection{LLMs} \label{subsec:examples_llms}
We aimed to include a broad selection of \texttt{LangChain}-compatible LLMs, including several open source variants, some GPT models and a Claude model. This experiment evaluates different model versions than the explanation experiment discussed earlier. Claude 3.7 sonnet, deepseek-reasoner-api, o1 and o3-mini were accessed via official APIs, the remaining models were run locally. 
\begin{table}[H]
\centering
\caption{\label{tab:llms}LLMs used in MaRDI experiment}
\begin{tabular}{|l|l|l|}
\hline
LLM & number of parameters &version/hash \\\hline
claude-3-7-sonnet-20250219 & ?&20250219\\
deepseek-r1 & 70b & 0c1615a8ca32\\
deepseek-reasoner-api &?& date: 21.03.2025\\
exaone-deep & 32b & a9422f9a5071\\
gemma3 & 27b& a418f5838eaf\\
gpt-5.2-2025-12-11 &?& 2025-12-11\\
llama3.3 &70b& a6eb4748fd29\\
nemotron & 70b &  2262f047a28a\\
o1-2024-12-17 & ?& 2024-12-17\\
o3-2025-04-16 & ?& 2025-04-16\\
o3-mini-2025-01-31 &?& 2025-01-31\\
o4-mini-2025-04-16 & ? & 2025-04-16\\
phi4 & 14b & ac896e5b8b34\\
qwq &32b& 009cb3f08d74\\
\hline
\end{tabular}
\end{table}

\subsection{Statistical Analysis}
For each LLM, we computed the mean accuracies and associated 95-percent Wilson confidence intervals of the production of correct \texttt{Cypher} queries on the first attempt and within the allowed sequence of one initial try plus up to two amendments. We then calculated the same accuracy measures on a per-question basis. This analysis was performed across nine (for MaRDI) or five (Hyena KG) natural-language questions and fourteen LLMs.

\subsection{Experimental Design}
For each question in the benchmark set, we provided the question answering pipeline described in \autoref{sec:framework} with the question, the graph schema and an instruction prompt (\autoref{subsec:gen_prompt} or \autoref{subsec:hyena_prompt}). If the generated query was not judged to be correct by the researcher, up to two natural language amendments were supplied for the framework to improve the query, along with an amendment prompt (\autoref{subsec:amend_prompt}). If the query was still not deemed to be correct after this, it was counted as a failure. This workflow was executed for each LLM mentioned in \autoref{subsec:examples_llms}.

\subsection{Real-Life Example 1: Question Answering on the MaRDI Knowledge Graph}\label{sec:exp2}
To assess real-life performance on a software-oriented Knowledge Graph, we tested 14 LLMs on nine natural-language questions over a software-centric slice of the MaRDI Knowledge Graph (CRAN packages, Zenodo software, cited papers, and authors with zbMATH IDs). The underlying subgraph connects software packages, publications, and authors through citation and authorship relationships, requiring the models to reason over multi-step links between research outputs and contributors. The question set is intentionally small to support consistent cross-model comparison and to highlight typical error patterns rather than to provide exhaustive coverage. We evaluate whether models can arrive at a correct \texttt{Cypher} query after the initial generation and up to two user-submitted amendments, and we analyze which question types lead to systematic difficulties. 

For instance, a sample interaction might be: 
\begin{verbatim}
    Natural Language Question: 
        Which are the ten authors that created the most software packages?
        
    Initially Generated Query:
\end{verbatim}
\vspace{-0.5em}
\begin{lstlisting}[language=cypher]
        MATCH (a:Author)-[:HAS_AUTHOR]->(s:SoftwarePackage) 
        RETURN a.name, COUNT(s) AS packageCount ORDER BY packageCount DESC LIMIT 10
\end{lstlisting}

\begin{verbatim}
    User Amendment Request: 
        The has_author relationship is the wrong way around.
        
    Amended Query:
\end{verbatim}
\vspace{-0.5em}
\begin{lstlisting}[language=cypher]
        MATCH (s:SoftwarePackage)-[:HAS_AUTHOR]->(a:Author) 
        RETURN a.name, COUNT(s) AS packageCount ORDER BY packageCount DESC LIMIT 10
\end{lstlisting}

\subsubsection{Data}
\begin{figure}[H]
    \centering
    \includegraphics[width=0.6\linewidth]{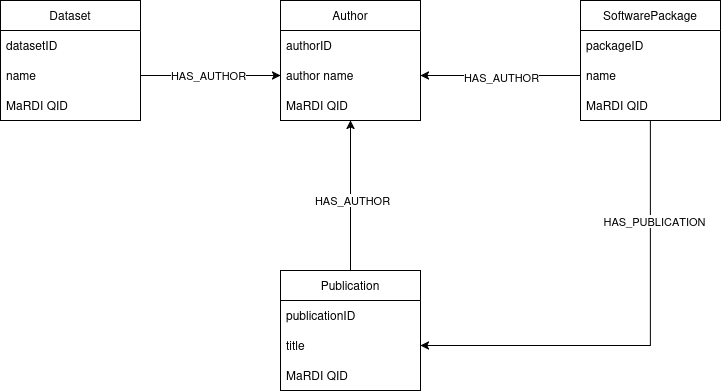}
    \caption{Data Model for the MaRDI KG subgraph}
    \label{fig:data_model}
\end{figure}

The data for this analysis consisted of a subset of the MaRDI Knowledge Graph and a custom set of natural-language questions based on the graph.
\subsubsubsection{Knowledge Graph}
We created a subgraph of the MaRDI Knowledge Graph \cite{schubotz2023bravo}, restricted to software packages, authors, and publications (\autoref{fig:data_model}). We extracted every CRAN package that cites a paper with a zbMATH de-number \cite{hornik2012comprehensive,petrera2021zbmath}, as well as the authors of those papers. We then added all developers who both maintain CRAN packages and hold a zbMATH author identifier. Finally, the subgraph was supplemented with every Zenodo software whose authors likewise possess a zbMATH author id \cite{peters2017zenodo}. The \texttt{SPARQL} queries used for these selections can be found in \autoref{sec:sparql}.

\subsubsubsection{Questions} \label{sec:mardi_questions}
We designed nine benchmark questions (see \autoref{sec:mardi_questions}) based on the graph schema of our MaRDI subgraph. Each can be translated into \texttt{Cypher} using various clauses such as \texttt{DISTINCT}, \texttt{LIMIT}, or \texttt{WHERE}. These clauses served only as guidelines: any query that returned the expected result was considered correct, regardless of the constructs used. Although this could in principle allow an incorrect query to match the expected output by coincidence, this is unlikely because the expected results are typically highly specific. An example question, which can be answered using the \texttt{LIMIT} clause, is \textit{"Which are the ten authors that created the most software packages?"}.

\subsubsection{Results}
\begin{figure}[H]
    \centering
    \includegraphics[width=0.85\linewidth]{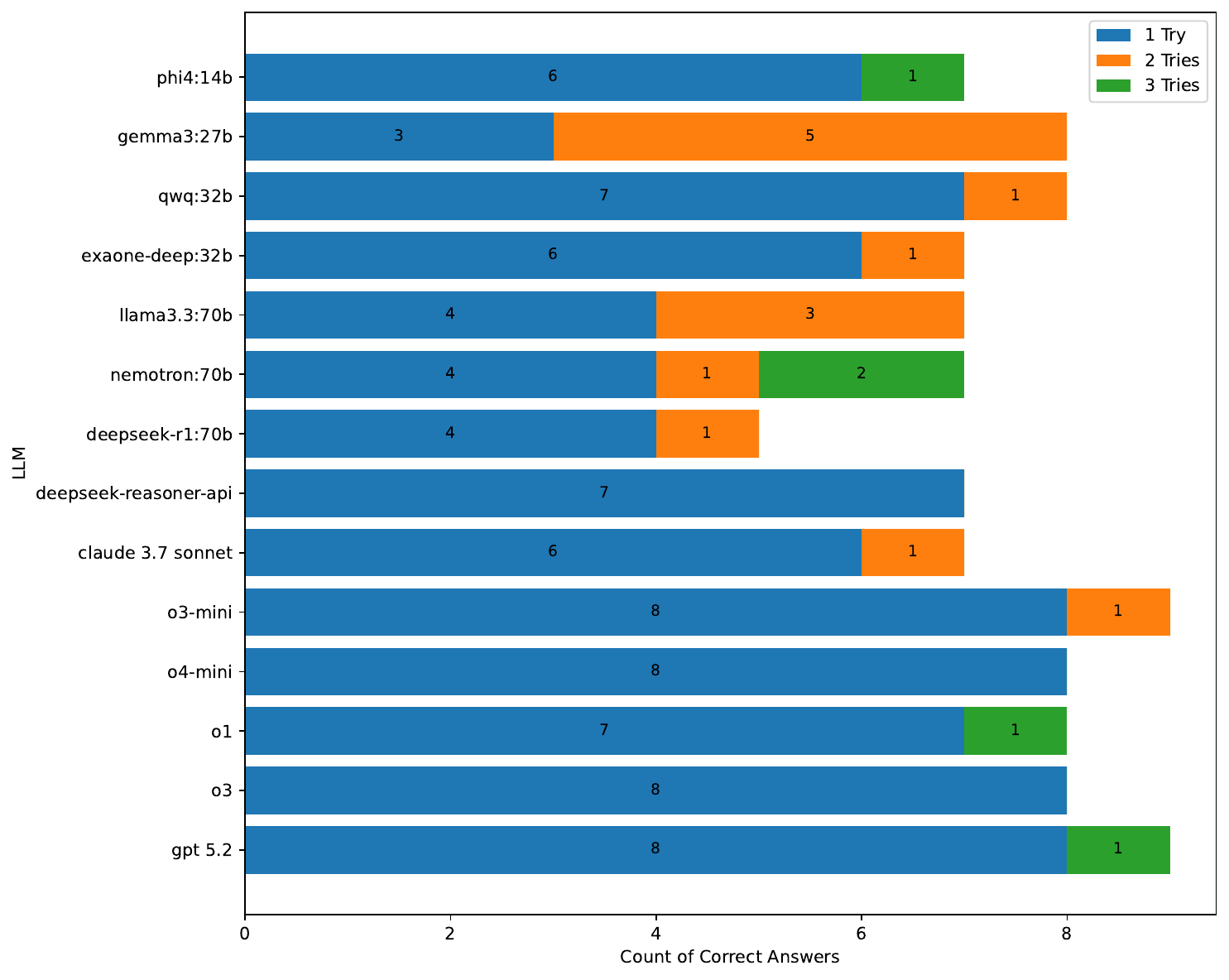}
    \caption{For each LLM, the bar length indicates the share of the nine questions eventually answered correctly. The colors within each bar partition that success rate by the number of attempts (first, second, or third) needed to reach the correct answer (a maximum of three tries was allowed).}
    \label{fig:mardi_correct_per_tries}
\end{figure}

We summarized the number of correct answers per model and tracked how many attempts the models needed for their correct answers in \autoref{fig:mardi_correct_per_tries}. The LLMs o3-mini and GPT 5.2 were the only ones that solved all nine tasks; qwq, gemma3, o4-mini, o1 and o3 followed with eight out of nine successfully solved tasks. Every other model reached 7/9 except deepseek-r1:70b, which managed only 5/9. The majority succeeded on the first try for the majority of their questions; only gemma3 more often required two attempts than one attempt, and a third attempt was seen only for GPT 5.2, o1, phi4:14b and nemotron:70b. If deepseek-reasoner-api, o4-mini and o3 were successful, it was always on the first try. The number of correct answers on the first try and within first tries as well as the accuracy and Wilson 95\% CI are listed in \autoref{tab:mardi_acc_9}.

\begin{figure}[H]
    \centering
    \includegraphics[width=0.85\linewidth]{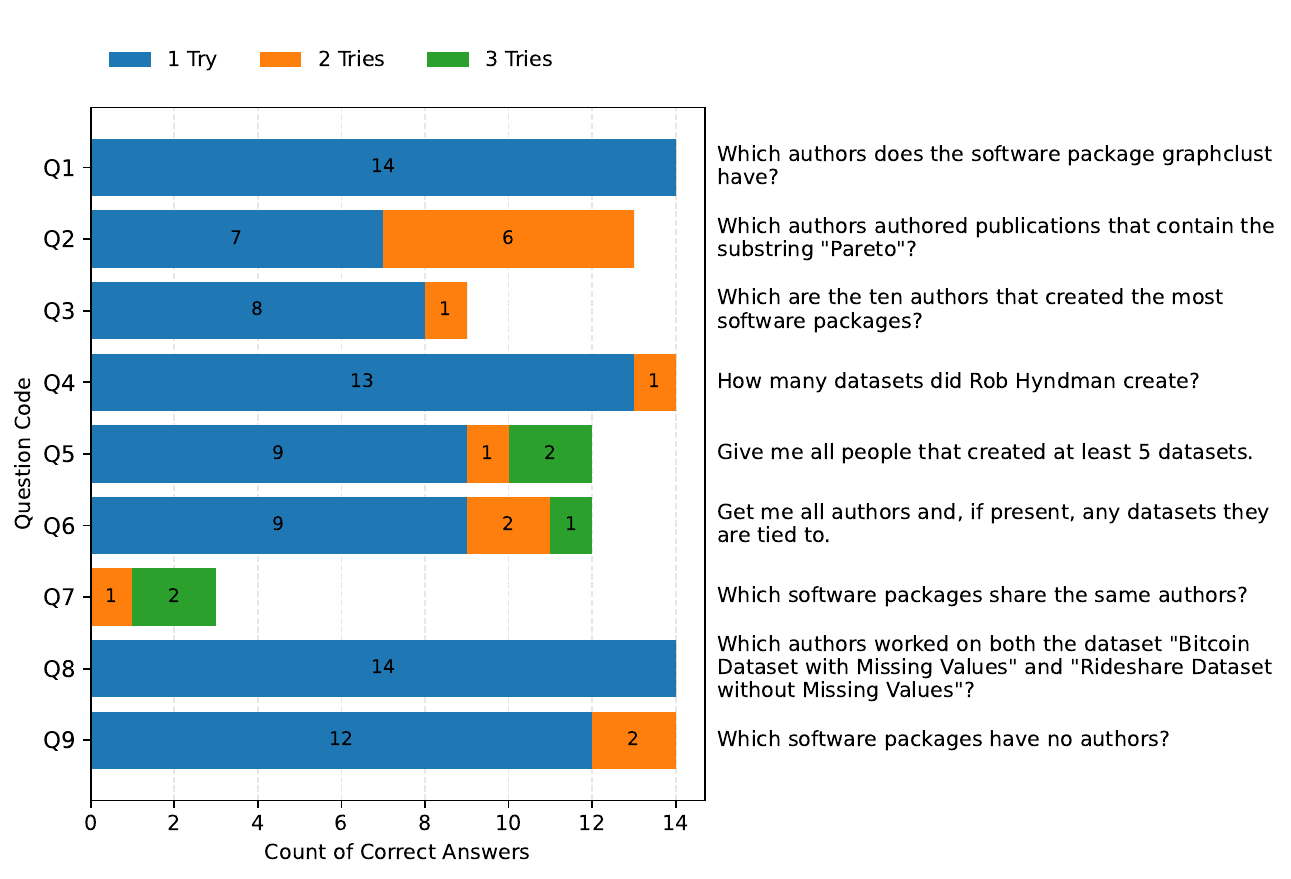}
    \caption{For each question, the bar height denotes the number of the 14 evaluated LLMs that returned the correct answer, illustrating the relative difficulty of the questions. Additionally, the bars are colored according to the number of amendment tries to solve them.}
    \label{fig:mardi_correct_per_question}
\end{figure}

\autoref{fig:mardi_correct_per_question} and \autoref{tab:question_accuracy_mardi} show the per-question accuracy (14 LLMs = maximum score 14), with the latter also containing total counts and CIs. Four tasks were successfully solved by every LLM (the count-based query, the “simple” query, the intersection query, and the \texttt{NOT} query), while the “shared authors between packages” task proved hardest, with just three correct solutions, and none of them on the first try.

\begin{table}[h]
\centering
\caption{First-attempt and within-three-attempts accuracy with 95\% Wilson CI on nine natural language questions.}
\label{tab:mardi_acc_9}
\resizebox{\textwidth}{!}{
\begin{tabular}{l r r l r r l}
\hline
LLM & correct on first try & accuracy first try & 95\% CI first try &
      correct within 3 & accuracy within 3 & 95\% CI within 3 \\
\hline
claude 3.7 sonnet        & 6 & 0.6667 & [0.354, 0.879] & 7 & 0.7778 & [0.453, 0.937] \\
deepseek-reasoner-api    & 7 & 0.7778 & [0.453, 0.937] & 7 & 0.7778 & [0.453, 0.937] \\
deepseek-r1:70b          & 4 & 0.4444 & [0.189, 0.733] & 5 & 0.5556 & [0.267, 0.811] \\
exaone\_deep:32b         & 6 & 0.6667 & [0.354, 0.879] & 7 & 0.7778 & [0.453, 0.937] \\
gemma3:27b               & 3 & 0.3333 & [0.121, 0.646] & 8 & 0.8889 & [0.565, 0.980] \\
gpt-5.2                  & 8 & 0.8889 & [0.565, 0.980] & 9 & 1.0000 & [0.701, 1.000] \\
llama3.3:70b             & 4 & 0.4444 & [0.189, 0.733] & 7 & 0.7778 & [0.453, 0.937] \\
nemotron:70b             & 4 & 0.4444 & [0.189, 0.733] & 7 & 0.7778 & [0.453, 0.937] \\
o1                       & 7 & 0.7778 & [0.453, 0.937] & 8 & 0.8889 & [0.565, 0.980] \\
o3-mini                  & 8 & 0.8889 & [0.565, 0.980] & 9 & 1.0000 & [0.701, 1.000] \\
o3                       & 8 & 0.8889 & [0.565, 0.980] & 8 & 0.8889 & [0.565, 0.980] \\
o4-mini                  & 8 & 0.8889 & [0.565, 0.980] & 8 & 0.8889 & [0.565, 0.980] \\
phi4:14b                 & 6 & 0.6667 & [0.354, 0.879] & 7 & 0.7778 & [0.453, 0.937] \\
qwq:32b                  & 7 & 0.7778 & [0.453, 0.937] & 8 & 0.8889 & [0.565, 0.980] \\
\hline
\end{tabular}
}
\end{table}

\begin{table}[h]
\centering
\small
\setlength{\tabcolsep}{4pt}
\renewcommand{\arraystretch}{1.1}
\caption{First-attempt and within-three-attempts accuracy across 14 models for each natural language question with 95\% Wilson CI.}
\label{tab:question_accuracy_mardi}
\resizebox{\textwidth}{!}{%
\begin{tabular}{p{0.45\textwidth} r r l r r l}
\hline
question &
first try correct &
first try accuracy &
95\% CI first try &
within 3 correct &
within 3 accuracy &
95\% CI within 3 \\
\hline
Get me all authors and, if present, any datasets they are tied to. &
9 & 0.642857 & [0.388, 0.837] & 12 & 0.857143 & [0.601, 0.960] \\
Give me all people that created at least 5 datasets. &
9 & 0.642857 & [0.388, 0.837] & 12 & 0.857143 & [0.601, 0.960] \\
How many datasets did Rob Hyndman create? &
13 & 0.928571 & [0.685, 0.987] & 14 & 1.000000 & [0.785, 1.000] \\
Which are the ten authors that created the most software packages? &
8 & 0.571429 & [0.326, 0.786] & 9 & 0.642857 & [0.388, 0.837] \\
Which authors authored publications that contain the substring ``Pareto''? &
7 & 0.500000 & [0.268, 0.732] & 13 & 0.928571 & [0.685, 0.987] \\
Which authors does the software package graphclust have? &
14 & 1.000000 & [0.785, 1.000] & 14 & 1.000000 & [0.785, 1.000] \\
Which authors worked on both the dataset ``Bitcoin Dataset with Missing Values'' and ``Rideshare Dataset without Missing Values''? &
14 & 1.000000 & [0.785, 1.000] & 14 & 1.000000 & [0.785, 1.000] \\
Which software packages have no authors? &
12 & 0.857143 & [0.601, 0.960] & 14 & 1.000000 & [0.785, 1.000] \\
Which software packages share the same authors? &
0 & 0.000000 & [0.000, 0.215] & 3 & 0.214286 & [0.076, 0.476] \\
\hline
\end{tabular}%
}
\end{table}

\subsection{Real-Life Example 2: Question Answering on a Hyena Knowledge Graph}\label{sec:exp3}
To determine whether the performance obtained for the KG used in the previous section remains similar for a real-world Knowledge Graph from a substantially different domain, we created a Knowledge Graph based on data from a long-term study of a free-ranging population of spotted hyenas inhabiting the Ngorongoro Crater in Tanzania (see \url{https://hyena-project.com}). We examined the performance of our approach for five real-world questions about kinship, which were directly formulated by experts involved in the Ngorongoro Hyena Project. We executed the query generation framework with 14 \texttt{LangChain}-compatible LLMs (see \autoref{tab:llms}). As in \autoref{sec:exp2}, we kept the number of questions low to allow systematic comparison across many models rather than exhaustive coverage of the graph. This setup does not identify the best-performing LLMs but highlights which ones fail or struggle under consistent conditions. The main research question we focus on is, again, whether the LLMs are able to produce a correct query after an initial \texttt{Cypher} query generation step and a maximum of two user-submitted amendments to the query. We also analyze the extent to which LLMs have more difficulties with certain questions.

A sample interaction might be:

\begin{verbatim}
    Natural Language Question:
        What is the proportion of cubs that were sired by philopatric males?
        
    Initially Generated Query:
\end{verbatim}  
 \vspace{-1em}
\begin{lstlisting}[language=cypher]   
        MATCH (cub:Hyena)-[:HAS_FATHER]->(dad:Hyena) 
        OPTIONAL MATCH (dad)-[:BIRTH_CLAN]->(bc:Clan) 
        OPTIONAL MATCH (dad)-[:CURRENT_CLAN]->(cc:Clan) 
        WITH COUNT(cub) AS totalCubs, 
        SUM(CASE WHEN dad.sex = 'male' AND bc = cc THEN 1 ELSE 0 END) AS philopatCubs 
        RETURN philopatCubs * 1.0 / totalCubs AS proportion
\end{lstlisting}
\begin{verbatim}
    User Amendment Request:
        With how this query is made, it is possible that the dad left, had the cub and came back. 
        Please check instead that he never changed his clan.

    Amended Query:
\end{verbatim}  
\vspace{-1em}
\begin{lstlisting}[language=cypher]
        MATCH (cub:Hyena)-[:HAS_FATHER]->(dad:Hyena) 
        OPTIONAL MATCH (dad)-[:BIRTH_CLAN]->(clan:Clan)<-[:CURRENT_CLAN]-(dad) 
        WITH COUNT(cub) AS totalCubs, 
        COUNT
            (CASE WHEN dad.sex = 'male' AND clan IS NOT NULL  
                AND NOT (dad)-[:CHANGED_CLAN]->() THEN 1 END) 
            AS philopatricCubs 
        RETURN toFloat(philopatricCubs) / totalCubs AS proportion
\end{lstlisting}

\subsubsection{Data}
This experiment is based on a knowledge graph constructed from long-term ecological observation data and a corresponding set of domain-expert natural-language questions.

\begin{figure}[H]
    \centering
    \includegraphics[width=0.5\linewidth]{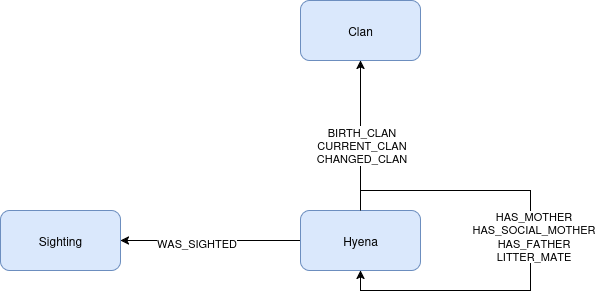}
    \caption{Data Model for the Hyena KG}
    \label{fig:hyena_kg}
\end{figure}

\subsubsubsection{Knowledge Graph}
We created a KG based on real observational data collected and curated by members of the Ngorongoro Hyena Project. To obtain a structurally meaningful yet manageable subset (see \autoref{fig:hyena_kg}), we focused on different kinship relationships between hyenas, such as parents, social mothers (since female spotted hyenas sometimes adopt a cub that is not their genetic offspring \cite{east2009maternal}) and litter mates (mothers produce between one and three cubs per delivery \cite{frank1991fatal}). Additionally, we integrated information about the birth clan and clans the hyenas belonged to at different points in time (most males change social group upon reaching sexual maturity \cite{honer2007female}), as well as sightings (where and when a given hyena was observed in the field).

\subsubsubsection{Questions}
We received realistic questions (\autoref{sec:hyena_questions}) from experts to be answered on the hyena graph. These five questions focus on the reproductive behaviour of males. The first two questions focus on paternities involving males that did not originate from the core population---a key biological variable influencing gene flow over time and space, associated with potential for local adaptation and risks of inbreeding depression. Those outsiders originate from clans outside the Ngorongoro Crater which are not directly monitored by the biologists, and are referred to as originating from clan X. The next two questions focus on how the dispersal status of males connects to their reproductive success. In spotted hyenas, male dispersal is thought to be a consequence of their reproductive access to females, with males remaining or leaving clans depending on mating opportunities \cite{honer2007female, davidian2016disperse}. The last question focuses on the extent to which males can manage to sire offspring in clans where they do not belong. Those events are thought to be rare since (i) intrusion into other clans can be met with resistance, (ii) spotted hyena males can only mate with females that accept them, and (iii) females are thought to prefer males they know well. Yet, it remains possible, especially if males visit a clan that they were previously a member of.

The hyena experts wrote the questions before the graph existed and thus were not familiar with our LLM/KG approach, thereby avoiding bias introduced by knowledge of the graph schema and mimicking a real-world trial of our framework. They selected them because they are representative of the kind of questions they routinely have to answer as part of larger analyses. The amendments to the \texttt{Cypher} queries generated by the LLMs, if necessary, were performed by the first author (L.P.) in accordance with detailed explanations of how the questions were understood by the biologists.

\subsubsection{Results}
\begin{figure}[H]
    \centering
    \includegraphics[width=0.85\linewidth]{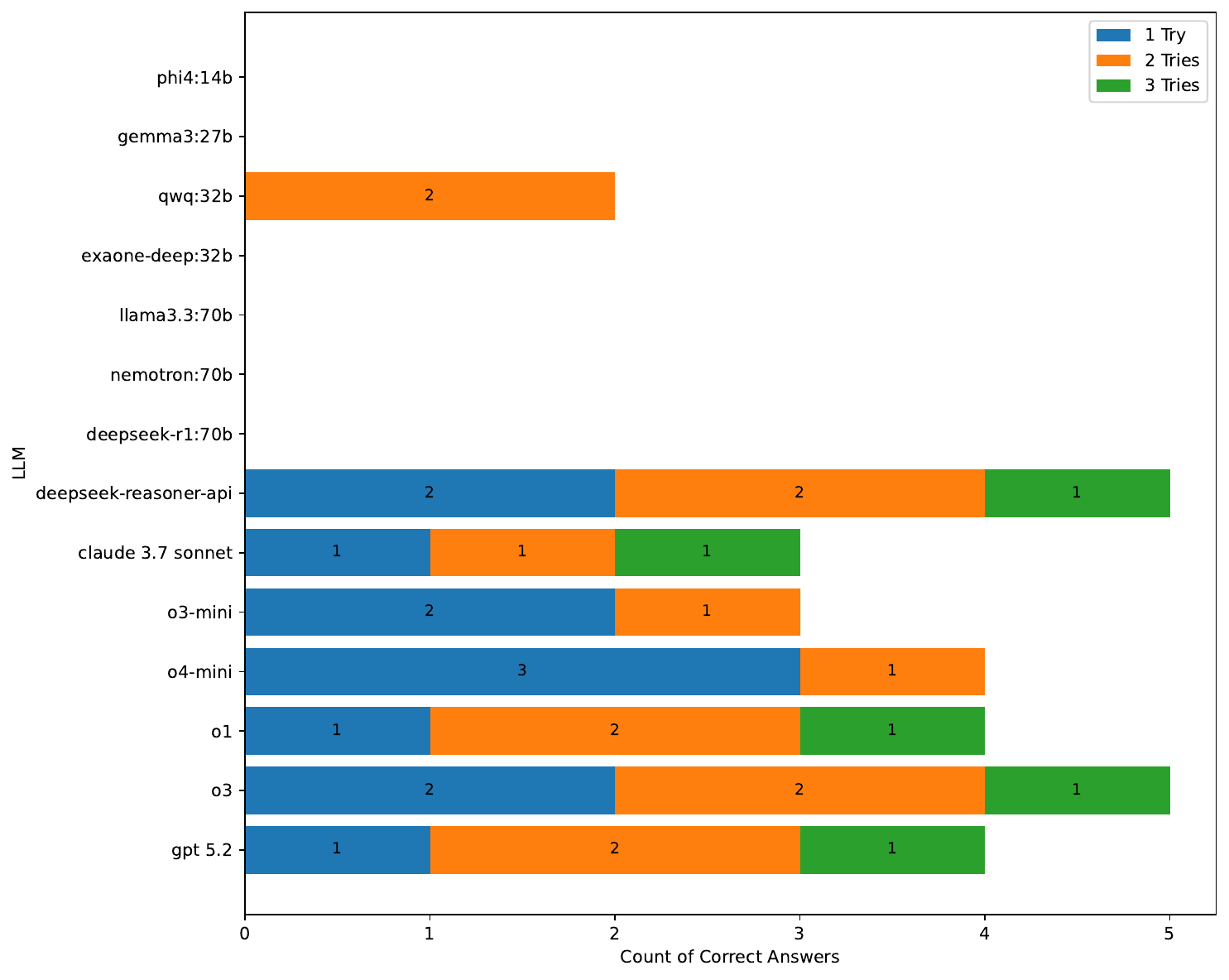}
    \caption{For each LLM, the bar height indicates the share of the five questions eventually answered correctly. The colors within each bar partition that success rate by the number of attempts (first, second, or third) needed to reach the correct answer (a maximum of three tries was allowed).}
    \label{fig:hyena_correct_tries}
\end{figure}

\begin{figure}[H]
    \centering
    \includegraphics[width=0.85\linewidth]{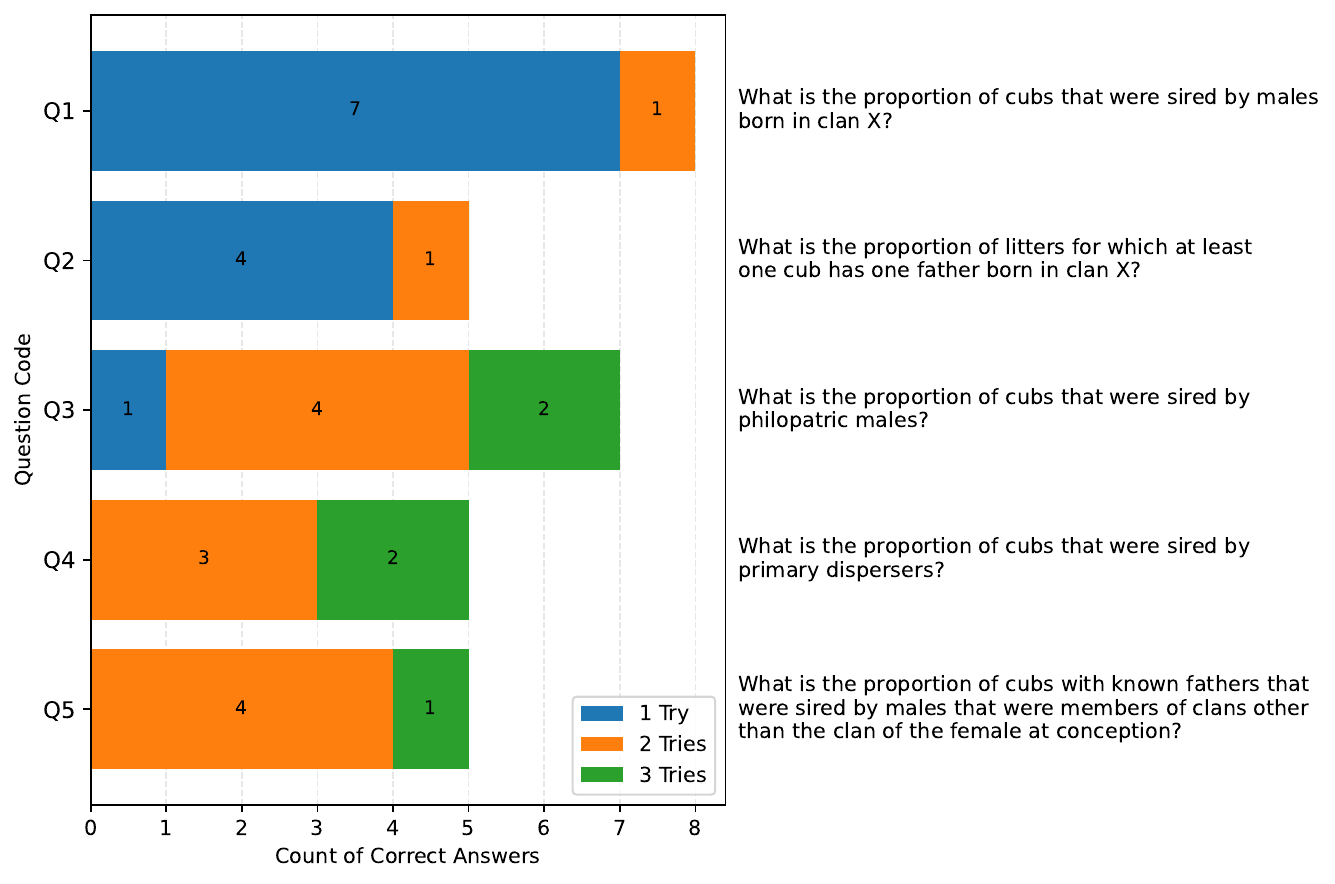}
    \caption{For each question, the bar length denotes the number of the 14 evaluated LLMs that returned the correct answer, illustrating the relative difficulty of the questions. Additionally, the bars are colored according to the number of amendment tries to solve them.}
    \label{fig:hyena_correct_question}
\end{figure}

Performance varied widely. Only o3 and deepseek-reasoner-api answered every question correctly (5/5). O4-mini, o1 and GPT 5.2 followed with four correct answers, and Claude and o3-mini solved three. Among the models run via Ollama, only qwq produced any correct responses (2/5); the remaining Ollama models failed to answer a single question. Even though the small sample introduces uncertainty, a 0/5 outcome still implies a low upper confidence limit of about 40 percent, so the poor performance of these models is unlikely to be an artifact of sample size.
Efficiency, measured as the number of attempts required per correct answer, was mixed. o4-mini and o3-mini provided most of their correct answers on the first try, whereas the other successful models needed at least as many second attempts as first ones, and more than half of them required three attempts for some questions (see \autoref{fig:hyena_correct_tries}). The number of correct answers on the first try and within first tries as well as the mean accuracy and Wilson 95\% CI are listed in \autoref{tab:hyena_acc}.
No single question was solved by every model, which is not surprising given that several models had zero correct answers overall (see \autoref{fig:hyena_correct_question} and \autoref{tab:hyena_question_accuracy}).

\begin{table}[h]
\centering
\caption{First-attempt and within-three-attempts accuracy with 95\% Wilson CI on five natural language questions.}
\label{tab:hyena_acc}
\resizebox{\textwidth}{!}{%
\begin{tabular}{l r r l r r l}
\hline
LLM & correct on first try & accuracy first try & 95\% CI first try &
      correct within 3 & accuracy within 3 & 95\% CI within 3 \\
\hline
claude 3.7 sonnet       & 1 & 0.2 & [0.036, 0.624] & 3 & 0.6 & [0.231, 0.882] \\
deepseek-reasoner-api   & 2 & 0.4 & [0.118, 0.769] & 5 & 1.0 & [0.566, 1.000] \\
deepseek-r1:70b         & 0 & 0.0 & [0.000, 0.434] & 0 & 0.0 & [0.000, 0.434] \\
exaone-deep:32b         & 0 & 0.0 & [0.000, 0.434] & 0 & 0.0 & [0.000, 0.434] \\
gemma3:27b              & 0 & 0.0 & [0.000, 0.434] & 0 & 0.0 & [0.000, 0.434] \\
gpt-5.2                 & 1 & 0.2 & [0.036, 0.624] & 4 & 0.8 & [0.376, 0.964] \\
llama3.3:70b            & 0 & 0.0 & [0.000, 0.434] & 0 & 0.0 & [0.000, 0.434] \\
nemotron:70b            & 0 & 0.0 & [0.000, 0.434] & 0 & 0.0 & [0.000, 0.434] \\
o1                      & 1 & 0.2 & [0.036, 0.624] & 4 & 0.8 & [0.376, 0.964] \\
o3                      & 2 & 0.4 & [0.118, 0.769] & 5 & 1.0 & [0.566, 1.000] \\
o3-mini                 & 2 & 0.4 & [0.118, 0.769] & 3 & 0.6 & [0.231, 0.882] \\
o4-mini                 & 3 & 0.6 & [0.231, 0.882] & 4 & 0.8 & [0.376, 0.964] \\
phi4:14b                & 0 & 0.0 & [0.000, 0.434] & 0 & 0.0 & [0.000, 0.434] \\
qwq:32b                 & 0 & 0.0 & [0.000, 0.434] & 2 & 0.4 & [0.118, 0.769] \\
\hline
\end{tabular}
}
\end{table}

\begin{table}[h]
\centering
\small
\setlength{\tabcolsep}{4pt}
\renewcommand{\arraystretch}{1.1}
\caption{First-attempt and within-three-attempts accuracy across 14 models for the hyena questions.}
\label{tab:hyena_question_accuracy}
\resizebox{\textwidth}{!}{%
\begin{tabular}{p{0.45\textwidth} r r l r r l}
\hline
question &
first try correct &
first try accuracy &
95\% CI first try &
within 3 correct &
within 3 accuracy &
95\% CI within 3 \\
\hline
What is the proportion of cubs that were sired by males born in clan X? &
7 & 0.500000 & [0.268, 0.732] & 8 & 0.571429 & [0.326, 0.786] \\
What is the proportion of cubs that were sired by philopatric males? &
1 & 0.071429 & [0.013, 0.315] & 7 & 0.500000 & [0.268, 0.732] \\
What is the proportion of cubs that were sired by primary dispersers? &
0 & 0.000000 & [0.000, 0.215] & 5 & 0.357143 & [0.163, 0.612] \\
What is the proportion of cubs with known fathers that were sired by males that were members of clans other than the clan of the female at conception? &
0 & 0.000000 & [0.000, 0.215] & 5 & 0.357143 & [0.163, 0.612] \\
What is the proportion of litters for which at least one cub has one father born in clan X? &
4 & 0.285714 & [0.117, 0.546] & 5 & 0.357143 & [0.163, 0.612] \\
\hline
\end{tabular}%
}
\end{table}

\subsection{Discussion}
Most large language models (LLMs) included in this experiment performed well on the questions in \autoref{sec:mardi_questions}, with two LLMs (o3-mini and GPT 5.2) generating all nine queries correctly and eleven others achieving seven or eight correct answers. Only one LLM, deepseek-r1:70b, generated only five correct queries. Allowing users to iteratively revise the query was still beneficial: while most correct queries were generated on the first attempt, a non-negligible share required a second or third amendment, especially for the models gemma3:27b, llama3.3:70b and nemotron. Our analysis further revealed which question structures were more difficult than others. For example, matching authors to a specific software package was handled reliably by every model, whereas identifying software packages which shared the same authors remained impossible for eleven of the fourteen LLMs.

However, strong performance on carefully designed questions did not transfer to the second, expert-driven case study in every case. When we applied the same pipeline to the Ngorongoro Hyena Knowledge Graph and five questions formulated by domain experts, performance gaps widened visibly. OpenAI o3 and deepseek-reasoner-api performed best (5/5 questions answered correctly), o4-mini, o1 and GPT 5.2 reached 4/5, Claude 3.7 Sonnet and o3-mini 3/5; among Ollama models, only qwq registered any correct answers (2/5), with the rest at 0. Efficiency also diverged: o4-mini most often succeeded on the first attempt, whereas other successful models mostly leaned on second or third tries. Had we relied on the MaRDI experiment alone, these differences would have remained hidden.

Two implications follow. First, expert-sourced questions are essential for discriminating among models that appear comparable on synthetic tests, and several Knowledge Graphs should be used to evaluate a method. Second, if a strategy with several amendment attempts is employed, efficiency, meaning how many tries a model needs, should be reported alongside the number of correct answers, because a rising number of retries makes the use of the tool more inconvenient.
In short, realistic questions and an additional Knowledge Graph revealed limitations of the pipeline using current LLMs, and future evaluations should not ignore the realism of questions and potential differences between domains.

\section{Conclusion}
We introduced an interactive framework that translates natural-language questions into auditable \texttt{Cypher}, explains the resulting queries, and lets users iteratively amend them. This pipeline brings transparency and control to KG-based question answering without asking users to learn a query language. Across three studies, we showed that LLMs can shoulder distinct roles in this pipeline with varying skill: they explain and sanity-check queries (\autoref{sec:exp1}), generate workable queries on a real mathematical KG (\autoref{sec:exp2}), and remain useful, though uneven, on expert-written questions in a biological KG (\autoref{sec:exp3}). 
On explanation quality, several models produced correct and complete one-sentence summaries for most movie-graph queries, yet a recurring failure was dropping explicit years - evidence that brevity pressures can harm faithfulness. These results form the most detailed and statistically grounded part of our evaluation.
Concerning query generation, many models solved the MaRDI tasks on the first attempt, while the Hyena study exposed wider gaps: only a few models consistently reached correct answers, and efficiency (attempts per success) varied markedly, underscoring why realistic, expert-sourced questions and cross-domain evaluation matter. These two studies function as targeted case evaluations of the framework on real KGs, illustrating how model behavior shifts with domain, data realism, and question style.
Future work could emphasize human-centered and evaluation-oriented directions: refining explanation prompts to reliably retain temporal and numeric constraints; designing stronger interactive tooling, such as diff views for query edits and side-by-side result previews to shorten the path from first draft to trusted query.
Taken together, our results support KG-centric, user-in-the-loop natural language QA systems with transparent, auditable result provenance.

\section{Acknowledgments}

We want to thank Oliver Höner \& Eve Davidian for providing us access to the hyena dataset and for formulating expert-level questions.  We thank all other members of the Ngorongoro Hyena Project which also collected data used here (Philemon Naman, Bettina Wachter \& Arjun Dheer). The authors also thank Leonie F. Walter for answering the hyena data questions independently using the \texttt{R} package \texttt{hyenaR} (without relying on LLM/KG). This work was supported by the Deutsche Forschungsgemeinschaft (DFG) (project grant 446 460135501).

\newpage

\printbibliography

\appendix 
\section{Prompts} \label{sec:prompts}

\subsection{Explanation Prompt} \label{subsec:explanation_prompt}
\begin{wrappedverb}
    You are a helpful assistant.
I have a Cypher query:

{query}

Please do these things:

    Provide a concise, step-by-step explanation of what this query does. Keep it as short as necessary to be clear and thorough, do not include unnecessary details.
    Conclude with a one- or two-sentence summary describing the query’s overall purpose.
    When trying to explain and summarize this query, please tell me if you noticed that something was wrong with the query, including:
    - relationships that are the wrong way around, for example: (b:Book)-[:READS]->(r:Scientist), where the correct one would be (b:Book)<-[:READS]-(r:Scientist)
    - relationship names that don´t make any sense in the context of the nodes they connect, for example: (r:Scientist)-[:SOLVES]->(b:Book); this is wrong because a book can't solve a scientist
    - a node name that makes no sense in the context: (r:Scientist)-[:READS]->(b:Airplane); this is wrong because you can't read an airplane
    - other things that you notice that don't make sense
\end{wrappedverb}

\subsection{Cypher Generation Prompt} \label{subsec:gen_prompt}
\begin{wrappedverb}
    Task:Generate Cypher statement to query a graph database.
Instructions:
Use only the provided relationship types and properties in the schema.
Do not use any other relationship types or properties that are not provided.
The cypher statement should only return nodes that are specifically asked for in the question.
Do not make the cypher query unnecessarily complex.
When the question asks for "What NODE_LABEL has X", the answer should be only the node name, not other details.
Cypher requires aggregate expressions, like COUNT(s), in the RETURN clause if you’re using them in the ORDER BY clause.
Example: MATCH (t:Tree)-[r:WAS_CUT]->(:Event) RETURN t.name, COUNT(*) AS cuttings ORDER BY cuttings
If it makes sense for the specific question and relationship, you can use bidirectional matching to matche the relationship in both directions.
Schema:
{schema}
Note: Do not include any explanations or apologies in your responses.
Do not respond to any questions that might ask anything else than for you to construct a Cypher statement.
Do not include any text except the generated Cypher statement.

The question is:
{question}
\end{wrappedverb}

\subsection{Amendment Prompt} \label{subsec:amend_prompt}
\begin{wrappedverb}
    You are given an existing Cypher query, a Knowledge Graph schema and an amendment request. Modify the Cypher query according to the amendment request.
The cypher query was generated on the basis of this question:
{question}
Existing Cypher query:
{current_query}
Graph schema:
{schema}
Amendment request:
{amendment}
Provide the updated Cypher query only, without any explanations.
\end{wrappedverb}

\subsection{Hyena Generation prompt}\label{subsec:hyena_prompt}
A line about how the sexes are entered into the graph is added in the prompt so the query does not fail because of different ways of writing this.
\begin{wrappedverb}
cypher_generation_template = """Task:Generate Cypher statement to query a graph database.
Instructions:
Use only the provided relationship types and properties in the schema.
Do not use any other relationship types or properties that are not provided.
The cypher statement should only return nodes that are specifically asked for in the question.
Do not make the cypher query unnecessarily complex.
When the question asks for "What NODE\_LABEL has X", the answer should be only the node name, not other details.
Cypher requires aggregate expressions, like COUNT(s), in the RETURN clause if you’re using them in the ORDER BY clause.
Example: MATCH (t:Tree)-[r:WAS_CUT]->(:Event) RETURN t.name, COUNT(*) AS cuttings ORDER BY cuttings
If it makes sense for the specific question and relationship, you can use bidirectional matching to matche the relationship in both directions.
The sexes in the graph are "male" and "female".
Schema:
{schema}
Note: Do not include any explanations or apologies in your responses.
Do not respond to any questions that might ask anything else than for you to construct a Cypher statement.
Do not include any text except the generated Cypher statement.

The question is:
{question}"""
\end{wrappedverb}

\section{SPARQL} \label{sec:sparql}
\begin{verbatim}
# get cran packages that cite papers that have a de nr (an identification number from zbMATH)
SELECT ?item ?itemLabel ?target ?targetLabel ?de_nr ?author ?authorLabel ?author_id 
WHERE
{
  ?item wdt:P31 wd:Q57080.
  ?item wdt:P223 ?target.
  ?target wdt:P1451 ?de_nr.
  ?target wdt:P16 ?author.
  ?author wdt:P676 ?author_id
  SERVICE wikibase:label { bd:serviceParam wikibase:language "en". }
}
\end{verbatim}

\begin{verbatim}
#get cran packages that have authors that have a zbMATH author id
SELECT ?item ?itemLabel ?target ?targetLabel ?author_id WHERE
{
  ?item wdt:P31 wd:Q57080.
  ?item wdt:P16 ?target.
  ?target wdt:P676 ?author_id
  SERVICE wikibase:label { bd:serviceParam wikibase:language "en". }
}
\end{verbatim}

\begin{verbatim}
#get zenodo packages that have authors with a zbMATH author id
SELECT ?item ?itemLabel ?target ?targetLabel ?author_id WHERE
{
  ?item wdt:P227 ?b.
  ?item wdt:P16 ?target.
  ?target wdt:P676 ?author_id
  SERVICE wikibase:label { bd:serviceParam wikibase:language "en". }
}
\end{verbatim}

\section{Explanation Queries}\label{sec:explanation_queries}
\begin{itemize}
  \item \textbf{Query:} \texttt{Cypher} query to be explained
  \item \textbf{N Hops:} Number of relationships between nodes in query
  \item \textbf{Clause type:} Types of clauses included in the query
  \item \textbf{Perturbation:} Type of error introduced into the query
\end{itemize}

\begingroup
\renewcommand{\arraystretch}{1.2} 
\emergencystretch=3em
\begin{longtable}{|p{10cm}|p{0.8cm}|p{1.5cm}|p{2.4cm}|}
\hline
Query & N Hops & Clause Type & Perturbation \\
\hline
\endfirsthead

\hline
Query & N Hops & Clause Type & Perturbation \\
\hline
\endhead
\code{MATCH (p:Person {name: "Alice"})-[:ACTED_IN]-> (m:Movie) RETURN p, m} & 1	& simple	& None\\
\code{MATCH (p:Person {name: "Alice"})<-[:ACTED_IN]- (m:Movie) RETURN p, m} &	1	& simple	& flip relationship\\
\code{MATCH (p:Person {name: "Alice"})-[:LIKES_TO_EAT]-> (m:Movie) RETURN p, m} &	1	& simple	& nonsense relation name\\
\code{MATCH (p:Person {name: "Alice"})-[:ACTED_IN]->(m:Food) RETURN p, m} &	1	& simple	& nonsense node name\\
\code{MATCH (celebrity {name: "Alice"})-[:ACTED_IN]-> (m:Movie) RETURN celebrity, m} &	1	& simple	& no node label, misleading name\\
\code{MATCH (p:Person {name: "Alice"})-[:ACTED_IN]-> (m:Movie)<-[:HAS_FAVORITE]-(c:Critic) RETURN p, m, c} &	2	& simple	& None\\
\code{MATCH (p:Person {name: "Alice"})<-[:ACTED_IN]- (m:Movie)<-[:HAS_FAVORITE]-(c:Critic) RETURN p, m, c} &	2	& simple	& flip relationship\\
\code{MATCH (p:Person {name: "Alice"})-[:LIKES_TO_EAT]-> (m:Movie)<-[:HAS_FAVORITE]-(c:Critic) RETURN p, m, c} &	2	& simple	& nonsense relation name\\
\code{MATCH (p:Person {name: "Alice"})-[:ACTED_IN]-> (m:Food)<-[:HAS_FAVORITE]-(c:Critic) RETURN p, m, c} &	2	& simple	& nonsense node name\\
\code{MATCH (celebrity {name: "Alice"})-[:ACTED_IN]-> (m:Movie)<-[:HAS_FAVORITE]-(c:Critic) RETURN celebrity, m, c} &	2	& simple	& no node label, misleading name\\
\code{MATCH (p:Person {name: "Alice"})-[:ACTED_IN]-> (m:Movie)<-[:HAS_FAVORITE]-(c:Critic)-[:BIRTH_CITY]-> (ct:City) RETURN p, m, c, ct} &	3	& simple	& None\\
\code{MATCH (p:Person {name: "Alice"})<-[:ACTED_IN]- (m:Movie)<-[:HAS_FAVORITE]-(c:Critic)-[:BIRTH_CITY]-> (ct:City) RETURN p, m, c, ct} &	3	& simple	& flip relationship\\
\code{MATCH (p:Person {name: "Alice"})-[:LIKES_TO_EAT]-> (m:Movie)<-[:HAS_FAVORITE]-(c:Critic)-[:BIRTH_CITY]-> (ct:City) RETURN p, m, c, ct} &	3	& simple	& nonsense relation name\\
\code{MATCH (p:Person {name: "Alice"})-[:ACTED_IN]->(m:Food) <-[:HAS_FAVORITE]-(c:Critic)-[:BIRTH_CITY]->(ct:City) RETURN p, m, c, ct} &	3	& simple	& nonsense node name\\
\code{MATCH (celebrity {name: "Alice"})-[:ACTED_IN]-> (m:Movie)<-[:HAS_FAVORITE]- (c:Critic) -[:BIRTH_CITY]-> (ct:City) RETURN celebrity, m, c, ct} &	3	& simple	& no node label, misleading name\\
\code{MATCH (p:Person {name: "Alice"})-[:ACTED_IN]-> (m:Movie) OPTIONAL MATCH (p)-[:DIRECTED]->(m2:Movie) RETURN p, m, m2} &	1 (+1)	& optional match	& None\\
\code{MATCH (p:Person {name: "Alice"})<-[:ACTED_IN] -(m:Movie) OPTIONAL MATCH (p)-[:DIRECTED]->(m2:Movie) RETURN p, m, m2} &	1 (+1)	& optional match	& flip relationship\\
\code{MATCH (p:Person {name: "Alice"})-[:LIKES_TO_EAT] ->(m:Movie) OPTIONAL MATCH (p)-[:DIRECTED]->(m2:Movie) RETURN p, m, m2} &	1 (+1)	& optional match	& nonsense relation name\\
\code{MATCH (p:Person {name: "Alice"})-[:ACTED_IN]->(m:Food) OPTIONAL MATCH (p)-[:DIRECTED]->(m2:Movie) RETURN p, m, m2} &	1 (+1)	& optional match	& nonsense node name\\
\code{MATCH (celebrity {name: "Alice"})-[:ACTED_IN] ->(m:Movie) OPTIONAL MATCH (celebrity)-[:DIRECTED] ->(m2:Movie) RETURN celebrity, m, m2} &	1 (+1)	& optional match	& no node label, misleading name\\
\code{MATCH (p:Person {name: "Alice"})-[:ACTED_IN] ->(m:Movie)<-[:HAS_FAVORITE]-(c:Critic) OPTIONAL MATCH (p)-[:DIRECTED]->(m2:Movie) RETURN p, m, c, m2} &	2 (+1)	& optional match	& None\\
\code{MATCH (p:Person {name: "Alice"})<-[:ACTED_IN] -(m:Movie)<-[:HAS_FAVORITE]-(c:Critic) OPTIONAL MATCH (p)-[:DIRECTED]->(m2:Movie) RETURN p, m, c, m2} &	2 (+1)	& optional match	& flip relationship\\
\code{MATCH (p:Person {name: "Alice"})-[:LIKES_TO_EAT] ->(m:Movie)<-[:HAS_FAVORITE]-(c:Critic) OPTIONAL MATCH (p)-[:DIRECTED]->(m2:Movie) RETURN p, m, c, m2} &	2 (+1)	& optional match	& nonsense relation name\\
\code{MATCH (p:Person {name: "Alice"})-[:ACTED_IN]->(m:Food) <-[:HAS_FAVORITE]-(c:Critic) OPTIONAL MATCH (p)-[:DIRECTED]->(m2:Movie) RETURN p, m, c, m2} &	2 (+1)	& optional match	& nonsense node name\\
\code{MATCH (celebrity {name: "Alice"})-[:ACTED_IN]-> (m:Movie)<-[:HAS_FAVORITE]-(c:Critic) OPTIONAL MATCH (celebrity)-[:DIRECTED]->(m2:Movie) RETURN celebrity, m, c, m2} &	2 (+1)	& optional match	& no node label, misleading name\\
\code{MATCH (p:Person {name: "Alice"})-[:ACTED_IN]-> (m:Movie)<-[:HAS_FAVORITE]-(c:Critic)-[:BIRTH_CITY] ->(ct:City) OPTIONAL MATCH (p)-[:DIRECTED]->(m2:Movie) RETURN p, m, c, ct, m2} &	3 (+1)	& optional match	& None\\
\code{MATCH (p:Person {name: "Alice"})<-[:ACTED_IN]- (m:Movie)<-[:HAS_FAVORITE]-(c:Critic)-[:BIRTH_CITY] ->(ct:City) OPTIONAL MATCH (p)-[:DIRECTED]->(m2:Movie) RETURN p, m, c, ct, m2} &	3 (+1)	& optional match	& flip relationship\\
\code{MATCH (p:Person {name: "Alice"})-[:LIKES_TO_EAT]-> (m:Movie)<-[:HAS_FAVORITE]-(c:Critic)-[:BIRTH_CITY]-> (ct:City) OPTIONAL MATCH (p)-[:DIRECTED]->(m2:Movie) RETURN p, m, c, ct, m2} &	3 (+1)	& optional match	& nonsense relation name\\
\code{MATCH (p:Person {name: "Alice"})-[:ACTED_IN]->(m:Food) <-[:HAS_FAVORITE]-(c:Critic)-[:BIRTH_CITY] ->(ct:City) OPTIONAL MATCH (p)-[:DIRECTED]->(m2:Movie) RETURN p, m, c, ct, m2} &	3 (+1)	& optional match	& nonsense node name\\
\code{MATCH (celebrity {name: "Alice"})-[:ACTED_IN]-> (m:Movie)<-[:HAS_FAVORITE]-(c:Critic)-[:BIRTH_CITY]-> (ct:City) OPTIONAL MATCH (celebrity)-[:DIRECTED]-> (m2:Movie) RETURN celebrity, m, c, ct, m2} &	3 (+1)	& optional match	& no node label, misleading name\\
\code{MATCH (p:Person {name: "Alice"})-[:ACTED_IN]-> (m:Movie) WITH p, m MATCH (m)<-[:ACTED_IN]- (coActor:Person) RETURN p, m, collect(DISTINCT coActor) AS coactors} &	1 (+1)	& with clause	& None\\
\code{MATCH (p:Person {name: "Alice"})<-[:ACTED_IN]- (m:Movie) WITH p, m MATCH (m)<-[:ACTED_IN]- (coActor:Person) RETURN p, m, collect(DISTINCT coActor) AS coactors} &	1 (+1)	& with clause	& flip relationship\\
\code{MATCH (p:Person {name: "Alice"})-[:LIKES_TO_EAT]-> (m:Movie) WITH p, m MATCH (m)<-[:ACTED_IN]- (coActor:Person) RETURN p, m, collect(DISTINCT coActor) AS coactors} &	1 (+1)	& with clause	& nonsense relation name\\
\code{MATCH (p:Person {name: "Alice"})-[:ACTED_IN]->(m:Food) WITH p, m MATCH (m)<-[:ACTED_IN]- (coActor:Person) RETURN p, m, collect(DISTINCT coActor) AS coactors} &	1 (+1)	& with clause	& nonsense node name\\
\code{MATCH (celebrity {name: "Alice"})-[:ACTED_IN]-> (m:Movie) WITH celebrity, m MATCH (m)<-[:ACTED_IN]-(coActor:Person) RETURN celebrity, m, collect(DISTINCT coActor) AS coactors} &	1 (+1)	& with clause	& no node label, misleading name\\
\code{MATCH (p:Person {name: "Alice"})-[:ACTED_IN]-> (m:Movie)<-[:HAS_FAVORITE]-(c:Critic) WITH p, m,c MATCH (m)<-[:ACTED_IN]-(coActor:Person) RETURN p, m, c, collect(DISTINCT coActor) AS coactors} &	2 (+1)	& with clause	& None\\
\code{MATCH (p:Person {name: "Alice"})<-[:ACTED_IN]- (m:Movie)<-[:HAS_FAVORITE]-(c:Critic) WITH p, m, c MATCH (m)<-[:ACTED_IN]-(coActor:Person) RETURN p, m, c, collect(DISTINCT coActor) AS coactors} &	2 (+1)	& with clause	& flip relationship\\
\code{MATCH (p:Person {name: "Alice"})-[:LIKES_TO_EAT]-> (m:Movie)<-[:HAS_FAVORITE]-(c:Critic) WITH p, m, c MATCH (m)<-[:ACTED_IN]-(coActor:Person) RETURN p, m, c, collect(DISTINCT coActor) AS coactors} &	2 (+1)	& with clause	& nonsense relation name\\
\code{MATCH (p:Person {name: "Alice"})-[:ACTED_IN]->(m:Food) <-[:HAS_FAVORITE]-(c:Critic) WITH p, m, c MATCH (m)<-[:ACTED_IN]-(coActor:Person) RETURN p, m, c, collect(DISTINCT coActor) AS coactors} &	2 (+1)	& with clause	& nonsense node name\\
\code{MATCH (celebrity {name: "Alice"})-[:ACTED_IN]-> (m:Movie)<-[:HAS_FAVORITE]-(c:Critic) WITH celebrity, m, c MATCH (m)<-[:ACTED_IN]-(coActor:Person) RETURN celebrity, m, c, collect(DISTINCT coActor) AS coactors} &	2 (+1)	& with clause	& no node label, misleading name\\
\code{MATCH (p:Person {name: "Alice"})-[:ACTED_IN]-> (m:Movie)<-[:HAS_FAVORITE]-(c:Critic)-[:BIRTH_CITY]-> (ct:City) WITH p, m, c, ct MATCH (m)<-[:ACTED_IN]-(coActor:Person) RETURN p, m, c, ct, collect(DISTINCT coActor) AS coactors} &	3 (+1)	& with clause	& None\\
\code{MATCH (p:Person {name: "Alice"})<-[:ACTED_IN]- (m:Movie)<-[:HAS_FAVORITE]-(c:Critic)-[:BIRTH_CITY]-> (ct:City) WITH p, m, c, ct MATCH (m)<-[:ACTED_IN]- (coActor:Person) RETURN p, m, c, ct, collect(DISTINCT coActor) AS coactors} &	3 (+1)	& with clause	& flip relationship\\
\code{MATCH (p:Person {name: "Alice"})-[:LIKES_TO_EAT]-> (m:Movie)<-[:HAS_FAVORITE]-(c:Critic)-[:BIRTH_CITY]-> (ct:City) WITH p, m, c, ct MATCH (m)<-[:ACTED_IN]-(coActor:Person) RETURN p, m, c, ct, collect(DISTINCT coActor) AS coactors} &	3 (+1)	& with clause	& nonsense relation name\\
\code{MATCH (p:Person {name: "Alice"})-[:ACTED_IN]->(m:Food) <-[:HAS_FAVORITE]-(c:Critic)-[:BIRTH_CITY]->(ct:City) WITH p, m, c, ct MATCH (m)<-[:ACTED_IN]- (coActor:Person) RETURN p, m, c, ct, collect(DISTINCT coActor) AS coactors} &	3 (+1)	& with clause	& nonsense node name\\
\code{MATCH (celebrity {name: "Alice"})-[:ACTED_IN]-> (m:Movie)<-[:HAS_FAVORITE]-(c:Critic)-[:BIRTH_CITY]-> (ct:City) WITH celebrity, m, c, ct MATCH (m)<-[:ACTED_IN]-(coActor:Person) RETURN celebrity, m, c, ct, collect(DISTINCT coActor) AS coactors} &	3 (+1)	& with clause	& no node label, misleading name\\
\code{MATCH (p:Person {name: "Alice"})-[:ACTED_IN]-> (m:Movie) WHERE m.release_year > 2020 RETURN p, m} &	1	& where clause	& None\\
\code{MATCH (p:Person {name: "Alice"})<-[:ACTED_IN]- (m:Movie) WHERE m.release_year > 2020 RETURN p, m} &	1	& where clause	& flip relationship\\
\code{MATCH (p:Person {name: "Alice"})-[:LIKES_TO_EAT]-> (m:Movie) WHERE m.release_year > 2020 RETURN p, m} &	1	& where clause	& nonsense relation name\\
\code{MATCH (p:Person {name: "Alice"})-[:ACTED_IN]->(m:Food) WHERE m.release_year > 2020 RETURN p, m} &	1	& where clause	& nonsense node name\\
\code{MATCH (celebrity {name: "Alice"})-[:ACTED_IN]-> (m:Movie) WHERE m.release_year > 2020 RETURN celebrity, m} &	1	& where clause	& no node label, misleading name\\
\code{MATCH (p:Person {name: "Alice"})-[:ACTED_IN]-> (m:Movie) WHERE m.release_year = -1 RETURN p, m} &	1	& where clause	& illogical where value\\
\code{MATCH (p:Person {name: "Alice"})-[:ACTED_IN]-> (m:Movie) WHERE m.title > 2020 RETURN p, m} &	1	& where clause	& wrong where type\\
\code{MATCH (p:Person {name: "Alice"})-[:ACTED_IN]-> (m:Movie) WHERE m.release_year > 2020 AND m.release_year < 2010 RETURN p, m} &	1	& where clause	& contradictory where values\\
\code{MATCH (p:Person {name: "Alice"})-[:ACTED_IN]-> (m:Movie)<-[:HAS_FAVORITE]-(c:Critic) WHERE m.release_year > 2020 RETURN p, m, c} &	2	& where clause	& None\\
\code{MATCH (p:Person {name: "Alice"})<-[:ACTED_IN]- (m:Movie)<-[:HAS_FAVORITE]-(c:Critic) WHERE m.release_year > 2020 RETURN p, m, c} &	2	& where clause	& flip relationship\\
\code{MATCH (p:Person {name: "Alice"})-[:LIKES_TO_EAT]-> (m:Movie)<-[:HAS_FAVORITE]-(c:Critic) WHERE m.release_year > 2020 RETURN p, m, c} &	2	& where clause	& nonsense relation name\\
\code{MATCH (p:Person {name: "Alice"})-[:ACTED_IN]->(m:Food) <-[:HAS_FAVORITE]-(c:Critic) WHERE m.release_year > 2020 RETURN p, m, c} &	2	& where clause	& nonsense node name\\
\code{MATCH (celebrity {name: "Alice"})-[:ACTED_IN]-> (m:Movie)<-[:HAS_FAVORITE]-(c:Critic) WHERE m.release_year > 2020 RETURN celebrity, m, c} &	2	& where clause	& no node label, misleading name\\
\code{MATCH (p:Person {name: "Alice"})-[:ACTED_IN]-> (m:Movie)<-[:HAS_FAVORITE]-(c:Critic) WHERE m.release_year = -1 RETURN p, m, c} &	2	& where clause	& illogical where value\\
\code{MATCH (p:Person {name: "Alice"})-[:ACTED_IN]-> (m:Movie)<-[:HAS_FAVORITE]-(c:Critic) WHERE m.title > 2020 RETURN p, m, c} &	2	& where clause	& wrong where type\\
\code{MATCH (p:Person {name: "Alice"})-[:ACTED_IN]-> (m:Movie)<-[:HAS_FAVORITE]-(c:Critic) WHERE m.release_year > 2020 AND m.release_year < 2010 RETURN p, m, c} &	2	& where clause	& contradictory where values\\
\code{MATCH (p:Person {name: "Alice"})-[:ACTED_IN]-> (m:Movie)<-[:HAS_FAVORITE]-(c:Critic)-[:BIRTH_CITY]-> (ct:City) WHERE m.release_year > 2020 RETURN p, m, c, ct} &	3	& where clause	& None\\
\code{MATCH (p:Person {name: "Alice"})<-[:ACTED_IN]- (m:Movie)<-[:HAS_FAVORITE]-(c:Critic)-[:BIRTH_CITY]-> (ct:City) WHERE m.release_year > 2020 RETURN p, m, c, ct} &	3	& where clause	& flip relationship\\
\code{MATCH (p:Person {name: "Alice"})-[:LIKES_TO_EAT]-> (m:Movie)<-[:HAS_FAVORITE]-(c:Critic)-[:BIRTH_CITY]-> (ct:City) WHERE m.release_year > 2020 RETURN p, m, c, ct} &	3	& where clause	& nonsense relation name\\
\code{MATCH (p:Person {name: "Alice"})-[:ACTED_IN]->(m:Food) <-[:HAS_FAVORITE]-(c:Critic)-[:BIRTH_CITY]->(ct:City) WHERE m.release_year > 2020 RETURN p, m, c, ct} &	3	& where clause	& nonsense node name\\
\code{MATCH (celebrity {name: "Alice"})-[:ACTED_IN]-> (m:Movie)<-[:HAS_FAVORITE]-(c:Critic)-[:BIRTH_CITY]-> (ct:City) WHERE m.release_year > 2020 RETURN celebrity, m, c, ct} &	3	& where clause	& no node label, misleading name\\
\code{MATCH (p:Person {name: "Alice"})-[:ACTED_IN]-> (m:Movie)<-[:HAS_FAVORITE]-(c:Critic)-[:BIRTH_CITY]-> (ct:City) WHERE m.release_year = -1 RETURN p, m, c, ct} &	3	& where clause	& illogical where value\\
\code{MATCH (p:Person {name: "Alice"})-[:ACTED_IN]-> (m:Movie)<-[:HAS_FAVORITE]-(c:Critic)-[:BIRTH_CITY]-> (ct:City) WHERE m.title > 2020 RETURN p, m, c, ct} &	3	& where clause	& wrong where type\\
\code{MATCH (p:Person {name: "Alice"})-[:ACTED_IN]-> (m:Movie)<-[:HAS_FAVORITE]-(c:Critic)-[:BIRTH_CITY]-> (ct:City) WHERE m.release_year > 2020 AND m.release_year < 2010 RETURN p, m, c, ct} &	3	& where clause	& contradictory where values\\
\code{MATCH (p:Person {name: "Alice"})-[:ACTED_IN]-> (m:Movie) RETURN p, m, CASE WHEN m.release_year < 2010 THEN 'old' ELSE 'new' END as movie_age} &	1	& case	& None\\
\code{MATCH (p:Person {name: "Alice"})<-[:ACTED_IN]- (m:Movie) RETURN p, m, CASE WHEN m.release_year < 2010 THEN 'old' ELSE 'new' END as movie_age} &	1	& case	& flip relationship\\
\code{MATCH (p:Person {name: "Alice"})-[:LIKES_TO_EAT]-> (m:Movie) RETURN p, m, CASE WHEN m.release_year < 2010 THEN 'old' ELSE 'new' END as movie_age} &	1	& case	& nonsense relation name\\
\code{MATCH (p:Person {name: "Alice"})-[:ACTED_IN]->(m:Food) RETURN p, m, CASE WHEN m.release_year < 2010 THEN 'old' ELSE 'new' END as movie_age} &	1	& case	& nonsense node name\\
\code{MATCH (celebrity {name: "Alice"})-[:ACTED_IN]-> (m:Movie) RETURN celebrity, m, CASE WHEN m.release_year < 2010 THEN 'old' ELSE 'new' END as movie_age} &	1	& case	& no node label, misleading name\\
\code{MATCH (p:Person {name: "Alice"})-[:ACTED_IN]-> (m:Movie) RETURN p, m, CASE WHEN m.release_year = -1 THEN 'old' ELSE 'new' END as movie_age} &	1	& case	& illogical where value\\
\code{MATCH (p:Person {name: "Alice"})-[:ACTED_IN]-> (m:Movie) RETURN p, m, CASE WHEN m.title < 2010 THEN 'old' ELSE 'new' END as movie_age} &	1	& case	& wrong where type\\
\code{MATCH (p:Person {name: "Alice"})-[:ACTED_IN]-> (m:Movie)<-[:HAS_FAVORITE]-(c:Critic) RETURN p, m, c, CASE WHEN m.release_year < 2010 THEN 'old' ELSE 'new' END as movie_age} &	2	& case	& None\\
\code{MATCH (p:Person {name: "Alice"})<-[:ACTED_IN]- (m:Movie)<-[:HAS_FAVORITE]-(c:Critic) RETURN p, m, c, CASE WHEN m.release_year < 2010 THEN 'old' ELSE 'new' END as movie_age} &	2	& case	& flip relationship\\
\code{MATCH (p:Person {name: "Alice"})-[:LIKES_TO_EAT]-> (m:Movie)<-[:HAS_FAVORITE]-(c:Critic) RETURN p, m, c, CASE WHEN m.release_year < 2010 THEN 'old' ELSE 'new' END as movie_age} &	2	& case	& nonsense relation name\\
\code{MATCH (p:Person {name: "Alice"})-[:ACTED_IN]->(m:Food) <-[:HAS_FAVORITE]-(c:Critic) RETURN p, m, c, CASE WHEN m.release_year < 2010 THEN 'old' ELSE 'new' END as movie_age} &	2	& case	& nonsense node name\\
\code{MATCH (celebrity {name: "Alice"})-[:ACTED_IN]-> (m:Movie)<-[:HAS_FAVORITE]-(c:Critic) RETURN celebrity, m, c, CASE WHEN m.release_year < 2010 THEN 'old' ELSE 'new' END as movie_age} &	2	& case	& no node label, misleading name\\
\code{MATCH (p:Person {name: "Alice"})-[:ACTED_IN]-> (m:Movie)<-[:HAS_FAVORITE]-(c:Critic) RETURN p, m, c, CASE WHEN m.release_year = -1 THEN 'old' ELSE 'new' END as movie_age} &	2	& case	& illogical where value\\
\code{MATCH (p:Person {name: "Alice"})-[:ACTED_IN]-> (m:Movie)<-[:HAS_FAVORITE]-(c:Critic) RETURN p, m, c, CASE WHEN m.title < 2010 THEN 'old' ELSE 'new' END as movie_age} &	2	& case	& wrong where type\\
\code{MATCH (p:Person {name: "Alice"})-[:ACTED_IN]-> (m:Movie)<-[:HAS_FAVORITE]-(c:Critic)-[:BIRTH_CITY]-> (ct:City) RETURN p, m, c, ct, CASE WHEN m.release_year < 2010 THEN 'old' ELSE 'new' END as movie_age} &	3	& case	& None\\
\code{MATCH (p:Person {name: "Alice"})<-[:ACTED_IN]- (m:Movie)<-[:HAS_FAVORITE]-(c:Critic)-[:BIRTH_CITY]-> (ct:City) RETURN p, m, c, ct, CASE WHEN m.release_year < 2010 THEN 'old' ELSE 'new' END as movie_age} &	3	& case	& flip relationship\\
\code{MATCH (p:Person {name: "Alice"})-[:LIKES_TO_EAT]-> (m:Movie)<-[:HAS_FAVORITE]-(c:Critic)-[:BIRTH_CITY]-> (ct:City) RETURN p, m, c, ct, CASE WHEN m.release_year < 2010 THEN 'old' ELSE 'new' END as movie_age} &	3	& case	& nonsense relation name\\
\code{MATCH (p:Person {name: "Alice"})-[:ACTED_IN]->(m:Food) <-[:HAS_FAVORITE]-(c:Critic)-[:BIRTH_CITY]->(ct:City) RETURN p, m, c, ct, CASE WHEN m.release_year < 2010 THEN 'old' ELSE 'new' END as movie_age} &	3	& case	& nonsense node name\\
\code{MATCH (celebrity {name: "Alice"})-[:ACTED_IN]-> (m:Movie)<-[:HAS_FAVORITE]-(c:Critic)-[:BIRTH_CITY]-> (ct:City) RETURN celebrity, m, c, ct, CASE WHEN m.release_year < 2010 THEN 'old' ELSE 'new' END as movie_age} &	3	& case	& no node label, misleading name\\
\code{MATCH (p:Person {name: "Alice"})-[:ACTED_IN]-> (m:Movie)<-[:HAS_FAVORITE]-(c:Critic)-[:BIRTH_CITY]-> (ct:City) RETURN p, m, c, ct, CASE WHEN m.release_year = -1 THEN 'old' ELSE 'new' END as movie_age} &	3	& case	& illogical where value\\
\code{MATCH (p:Person {name: "Alice"})-[:ACTED_IN]-> (m:Movie)<-[:HAS_FAVORITE]-(c:Critic)-[:BIRTH_CITY]-> (ct:City) RETURN p, m, c, ct, CASE WHEN m.title < 2010 THEN 'old' ELSE 'new' END as movie_age} &	3	& case	& wrong where type\\
\hline
\caption{Explanation queries}
\end{longtable}
\endgroup

\section{MaRDI Experiment Questions} \label{sec:mardi_questions}
\begin{itemize}
  \item \textbf{Question:} Input Question
  \item \textbf{Type:} Query features that are included in the reference query
  \item \textbf{Reference \texttt{Cypher} Query:} \texttt{Cypher} query that outputs the desired results, which the output from the generated query can be compared to
\end{itemize}
\begin{table}[H]
\small
\renewcommand{\arraystretch}{1.2} 
\caption{MaRDI experiment questions, question types and example Cypher queries}
\begin{tabular}{|p{4cm}|p{2cm}|p{10cm}|}
\hline
Question & Type & Reference Cypher Query \\\hline
Which authors does the software package graphclust have? & simple & \code{MATCH (sp:SoftwarePackage {name:"graphclust"})-[:HAS_AUTHOR]
->(a:Author) RETURN a;}\\
Which authors authored publications that contain the substring "Pareto"? & where & \code{MATCH (p:Publication) WHERE p.title CONTAINS "Pareto" MATCH (p)-[:HAS_AUTHOR]->(a:Author) RETURN DISTINCT a.name AS authorName, a.authorId AS authorId}\\
Which are the ten authors that created the most software packages? & top n & \code{MATCH (a:Author)<-[:HAS_AUTHOR]-(s:SoftwarePackage) RETURN a.name AS authorName, a.authorId AS authorId, COUNT(s) AS packageCount ORDER BY packageCount DESC LIMIT 10} \\
How many datasets did Rob Hyndman create? &  counts & \code{MATCH (a:Author { name: "Rob Hyndman" })<-[:HAS_AUTHOR]-(d:Dataset) RETURN count(d) AS numberOfDatasets} \\
Give me all people that created at least 5 datasets. & distinct & \code{MATCH (p:Author)<-[:HAS_AUTHOR]-(d:Dataset) WITH p, COUNT(d) AS numberOfDatasets WHERE numberOfDatasets >= 5 RETURN p.name AS authorName,  numberOfDatasets ORDER BY numberOfDatasets DESC}\\
Get me all authors and, if present, any datasets they are tied to. & optional match &\code{MATCH (a:Author) OPTIONAL MATCH (a)<-[:HAS_AUTHOR]
-(d:Dataset) RETURN a.name AS authorName, a.authorId AS authorId, COLLECT(d) AS datasets	}\\
Which software packages share the same authors?& shared  & \code{MATCH (p:SoftwarePackage)-[:HAS_AUTHOR]->(a:Author) WITH p, COLLECT(DISTINCT a.authorId) AS authorIds WITH p, apoc.coll.sort(authorIds) AS sortedAuthorIds WITH sortedAuthorIds, COLLECT(p) AS packages WHERE SIZE(packages) > 1 RETURN sortedAuthorIds AS authorGroup, packages}\\
Which authors worked on both the dataset "Bitcoin Dataset with Missing Values" and "Rideshare Dataset without Missing Values"?& intersection & \code{MATCH (a:Author)<-[:HAS_AUTHOR]-(d1:Dataset), (a)<-[:HAS_AUTHOR]-(d2:Dataset) WHERE d1.name = "Bitcoin Dataset with Missing Values" AND d2.name = "Rideshare Dataset without Missing Values" RETURN a.name AS authorName, a.authorId AS authorId}\\
Which software packages have no authors?& not & \code{MATCH (p:SoftwarePackage) WHERE NOT (p)-[:HAS_AUTHOR] ->(:Author) RETURN p.packageId AS packageId, p.name AS packageName}\\
\hline
\end{tabular}
\end{table}

\section{Hyena Experiment Questions}\label{sec:hyena_questions}
\begin{itemize}
  \item \textbf{Question:} Input Question
  \item \textbf{N Hops:} Number of relationships between nodes in the path based on the question
  \item \textbf{Includes Calculation:} If the question includes a calculation, such as a proportion
  \item \textbf{Includes Complicated Niche Concept:} If the question includes a concept which only an expert in the field would understand
\end{itemize}
\begin{table}[H]
\caption{Hyena experiment questions}
\begin{tabular}{|p{6cm}|p{2cm}|p{4cm}|p{4cm}|}
\hline
Question & N Hops  & Includes Calculation & Includes Complicated Niche Concept \\\hline
What is the proportion of cubs that were sired by males born in clan X? & 2 & True & False \\
What is the proportion of litters for which at least one cub has one father born in clan X? & 3 & True & True\\
What is the proportion of cubs that were sired by philopatric males? & 3 & True & True\\
What is the proportion of cubs that were sired by primary dispersers? & 3 & True & True\\
What is the proportion of cubs with known fathers that were sired by males that were members of clans other than the clan of the female at conception? & 6 & True & False\\
\hline
\end{tabular}
\end{table}

\section{Detailed Results}\label{sec:detailed_results}
\subsection{Explanation trials}\label{subsec:detailed_explanation_results}
The disagreements between models for the correct one-sentence summary are reported in \autoref{tab:explanation_first_detailed}. Disagreements related to correct perturbation detection appear in \autoref{tab:explanation_detection_detailed}, and disagreements for identifying false positives are summarized in \autoref{tab:false_positive_detailed}.

\begin{table}[h]
\centering
\caption{Total number of disagreements between the models in the explanation trial when generating one-sentence-summaries, further split up into the ones where only the row model or only the column model was correct}
\label{tab:explanation_first_detailed}
\begin{tabular}{l l r r r}
\hline
row\_model & col\_model & row\_only & col\_only & total\_diff \\
\hline
claude 3.7 sonnet & deepseek-reasoner-api & 7 & 26 & 33 \\
claude 3.7 sonnet & deepseek-r1:70b & 9 & 22 & 31 \\
claude 3.7 sonnet & o1-preview & 1 & 23 & 24 \\
claude 3.7 sonnet & o3-mini & 5 & 22 & 27 \\
deepseek-reasoner-api & deepseek-r1:70b & 15 & 9 & 24 \\
deepseek-reasoner-api & o1-preview & 7 & 10 & 17 \\
deepseek-reasoner-api & o3-mini & 12 & 10 & 22 \\
deepseek-r1:70b & o1-preview & 7 & 16 & 23 \\
deepseek-r1:70b & o3-mini & 9 & 13 & 22 \\
o1-preview & o3-mini & 13 & 8 & 21 \\
\hline
\end{tabular}
\end{table}

\begin{table}[h]
\centering
\caption{Total number of disagreements between the models in the additional trial, further split up into the ones where only the row model or only the column model was correct.}
\label{tab:explanation_detection_detailed}
\begin{tabular}{l l r r r}
\hline
row\_model & col\_model & row\_only & col\_only & total\_diff \\
\hline
claude 3.7 sonnet & deepseek-reasoner-api & 7 & 10 & 17 \\
claude 3.7 sonnet & deepseek-r1:70b       & 17 & 4  & 21 \\
claude 3.7 sonnet & o1-preview            & 6 & 8  & 14 \\
claude 3.7 sonnet & o3-mini               & 14 & 8 & 22 \\
deepseek-reasoner-api & deepseek-r1:70b   & 16 & 0 & 16 \\
deepseek-reasoner-api & o1-preview        & 3 & 2  & 5  \\
deepseek-reasoner-api & o3-mini           & 9 & 0  & 9  \\
deepseek-r1:70b & o1-preview              & 0 & 15 & 15 \\
deepseek-r1:70b & o3-mini                 & 1 & 8  & 9  \\
o1-preview & o3-mini                      & 8 & 0  & 8  \\
\hline
\end{tabular}
\end{table}

\begin{table}[h]
\centering
\caption{Total number of disagreements between the models in the false-positive experiment, split into cases where only the row model or only the column model was correct.}
\label{tab:false_positive_detailed}
\begin{tabular}{l l r r r}
\hline
row\_model & col\_model & row\_only & col\_only & total\_diff \\
\hline
claude 3.7 sonnet        & deepseek-reasoner-api & 0 & 5 & 5 \\
claude 3.7 sonnet        & deepseek-r1:70b       & 3 & 1 & 4 \\
claude 3.7 sonnet        & o1-preview            & 0 & 4 & 4 \\
claude 3.7 sonnet        & o3-mini               & 0 & 5 & 5 \\
deepseek-reasoner-api    & deepseek-r1:70b       & 7 & 0 & 7 \\
deepseek-reasoner-api    & o1-preview            & 1 & 0 & 1 \\
deepseek-reasoner-api    & o3-mini               & 0 & 0 & 0 \\
deepseek-r1:70b          & o1-preview            & 0 & 6 & 6 \\
deepseek-r1:70b          & o3-mini               & 0 & 7 & 7 \\
o1-preview               & o3-mini               & 0 & 1 & 1 \\
\hline
\end{tabular}
\end{table}

\end{document}